\documentclass[lettersize,journal]{IEEEtran}
\usepackage{amsmath,amsfonts}
\usepackage{algorithmic}
\usepackage{algorithm}
\usepackage{array}
\usepackage[caption=false,font=normalsize,labelfont=sf,textfont=sf]{subfig}
\usepackage{textcomp}
\usepackage{stfloats}
\usepackage{url}
\usepackage{verbatim}
\usepackage{graphicx}
\usepackage{cite}
\usepackage{booktabs}
\usepackage{orcidlink}
\usepackage{bm}
\usepackage{multirow}
\usepackage{float}
\usepackage{amstext}
\usepackage{amsthm}
\usepackage{mathrsfs}
\usepackage{makecell}
\usepackage{colortbl}
\usepackage{hyperref}
\usepackage{amssymb}
\title{\LARGE \bf Boosting Instance Awareness via Cross-View Correlation with 4D Radar and Camera for 3D Object Detection}

\author{Xiaokai Bai, Lianqing Zheng, Si-Yuan Cao, Xiaohan Zhang, Zhe Wu, Beinan Yu, Fang Wang, Jie Bai, and Hui-Liang Shen, \emph{Senior Member, IEEE}

\thanks{This work was supported in part by the National Key Research and Development Program of China under grant 2023YFB3209800. \emph{(Corresponding authors: Beinan Yu and Hui-Liang Shen.)}}

\thanks{Xiaokai Bai, Si-Yuan Cao, Xiaohan Zhang, Zhe Wu, and Hui-Liang Shen are with the College of Information Science and Electronic Engineering, Zhejiang University, Hangzhou 310027, China (e-mail: shawnnnkb@zju.edu.cn, cao\_siyuan@zju.edu.cn, zhangxh2023@zju.edu.cn, jeffw@zju.edu.cn, shenhl@zju.edu.cn).}

\thanks{Lianqing Zheng is with the School of Automotive Studies, Tongji University, Shanghai 201804, China (e-mail: zhenglianqing@tongji.edu.cn).}

\thanks{Beinan Yu is with the College of Computer Science and Technology, Zhejiang University, Hangzhou 310027, China, and also with the Jinhua Institute of Zhejiang University, Jinhua 321299, China (e-mail: mr\_vernon@hotmail.com)}

\thanks{Fang Wang and Jie Bai are with the School of Information and Electrical Engineering, Hangzhou City University, Hangzhou 310015, China, and also with the Hangzhou City University Binjiang Innovation Center, Hangzhou 310052, China (e-mail: wangf@zucc.edu.cn, baij@zucc.edu.cn).}

}% <-this % stops a space
\begin{document}

\markboth{Accepted by IEEE Transactions on Multimedia}
{Bai \MakeLowercase{\textit{et al.}}: Boosting Instance Awareness via Cross-View Correlation with 4D Radar and Camera for 3D Object Detection}
\maketitle

% \thispagestyle{empty}
% \pagestyle{empty}

%%%%%%%%%%%%%%%%%%%%%%%%%%%%%%%%%%%%%%%%%%%%%%%%%%%%%%%%%%%%%%%%%%%%%%%%%%%%%%%%
\begin{abstract}

4D millimeter-wave radar has emerged as a promising sensing modality for autonomous driving due to its robustness and affordability. However, its sparse and weak geometric cues make reliable instance activation difficult, limiting the effectiveness of existing radar–camera fusion paradigms. BEV-level fusion offers global scene understanding but suffers from weak instance focus, while perspective-level fusion captures instance details but lacks holistic context. To address these limitations, we propose SIFormer, a scene–instance aware transformer for 3D object detection using 4D radar and camera. SIFormer first suppresses background noise during view transformation through segmentation- and depth-guided localization. It then introduces a cross-view activation mechanism that injects 2D instance cues into BEV space, enabling reliable instance awareness under weak radar geometry. Finally, a transformer-based fusion module aggregates complementary image semantics and radar geometry for robust perception. As a result, with the aim of enhancing instance awareness, SIFormer bridges the gap between the two paradigms, combining their complementary strengths to address inherent sparse nature of radar and improve detection accuracy. Experiments demonstrate that SIFormer achieves state-of-the-art performance on View-of-Delft, TJ4DRadSet and NuScenes datasets. Source code is available at \href{https://github.com/shawnnnkb/SIFormer}{\textcolor[RGB]{213, 43, 107}{https://github.com/shawnnnkb/SIFormer}}.

\end{abstract}


%%%%%%%%%%%%%%%%%%%%%%%%%%%%%%%%%%%%%%%%%%%%%%%%%%%%%%%%%%%%%%%%%%%%%%%%%%%%%%%%
\begin{figure}[ht]
    \centering
    \includegraphics[width=\linewidth]{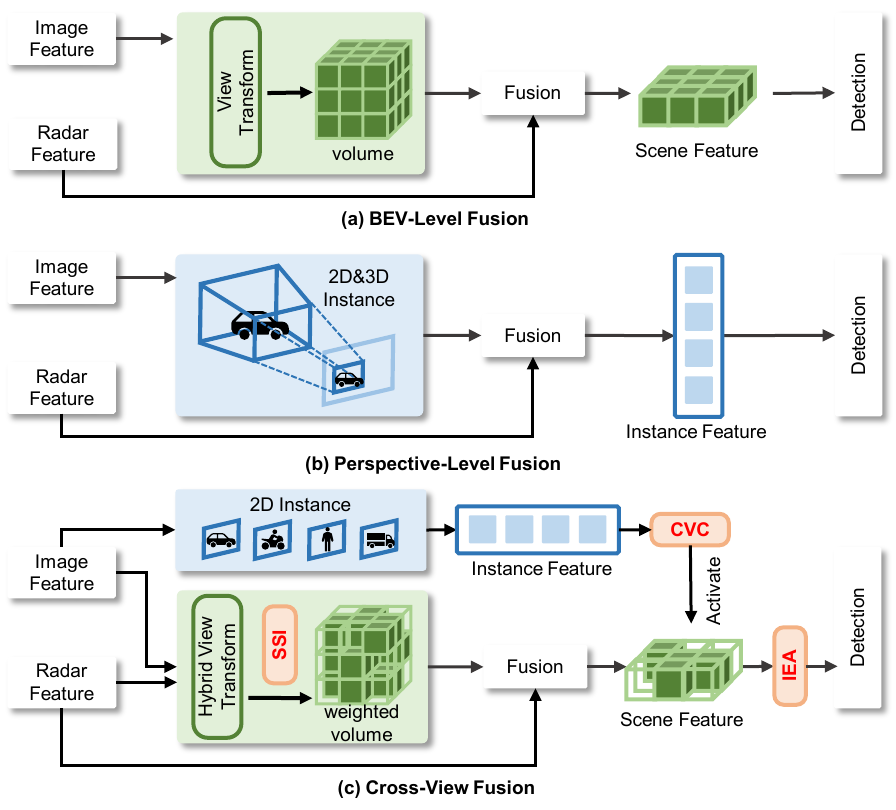}
    \caption{Comparison of radar and camera fusion pipelines. Previous radar-camera fusion models typically adopt either (a) BEV-level or (b) perspective-level fusion. Our SIFormer, with the aim of enhancing instance awareness, bridges the gap between the two paradigms while combining their complementary strengths, as shown in (c).}
    \label{fig:comparison}
\end{figure}

\section{Introduction}
Autonomous driving has become a prominent topic in academia and industry, focusing on perception, planning, and control \cite{Comprehensive_Survey}. Accurate perception is essential for safe and efficient driving, encompassing tasks like tracking, segmentation, semantic scene completion, and 3D object detection. The goal of 3D object detection is to classify and localize critical objects using sensor data, relying on cameras, LiDAR, and radar for complementary signals. Cameras provide high-resolution semantic information, capturing object color and texture, but lack depth perception \cite{TMMDAM}. LiDAR produces accurate 3D point clouds with strong geometric priors, but struggles with similarly shaped objects and lacks semantic cues like color and texture \cite{LIDARDET_review}. Both sensors can degrade in adverse weather \cite{badweather_degradation}. In contrast, 4D radar is more robust under poor lighting and weather conditions \cite{all-weather2}. With the introduction of 4D radar, which adds elevation and velocity information compared to conventional 3D radar, the 4D radar becomes capable of long-range detection with motion awareness \cite{VoD, TJ4D}. 

\begin{figure}[ht]
    \centering
    \includegraphics[width=\linewidth]{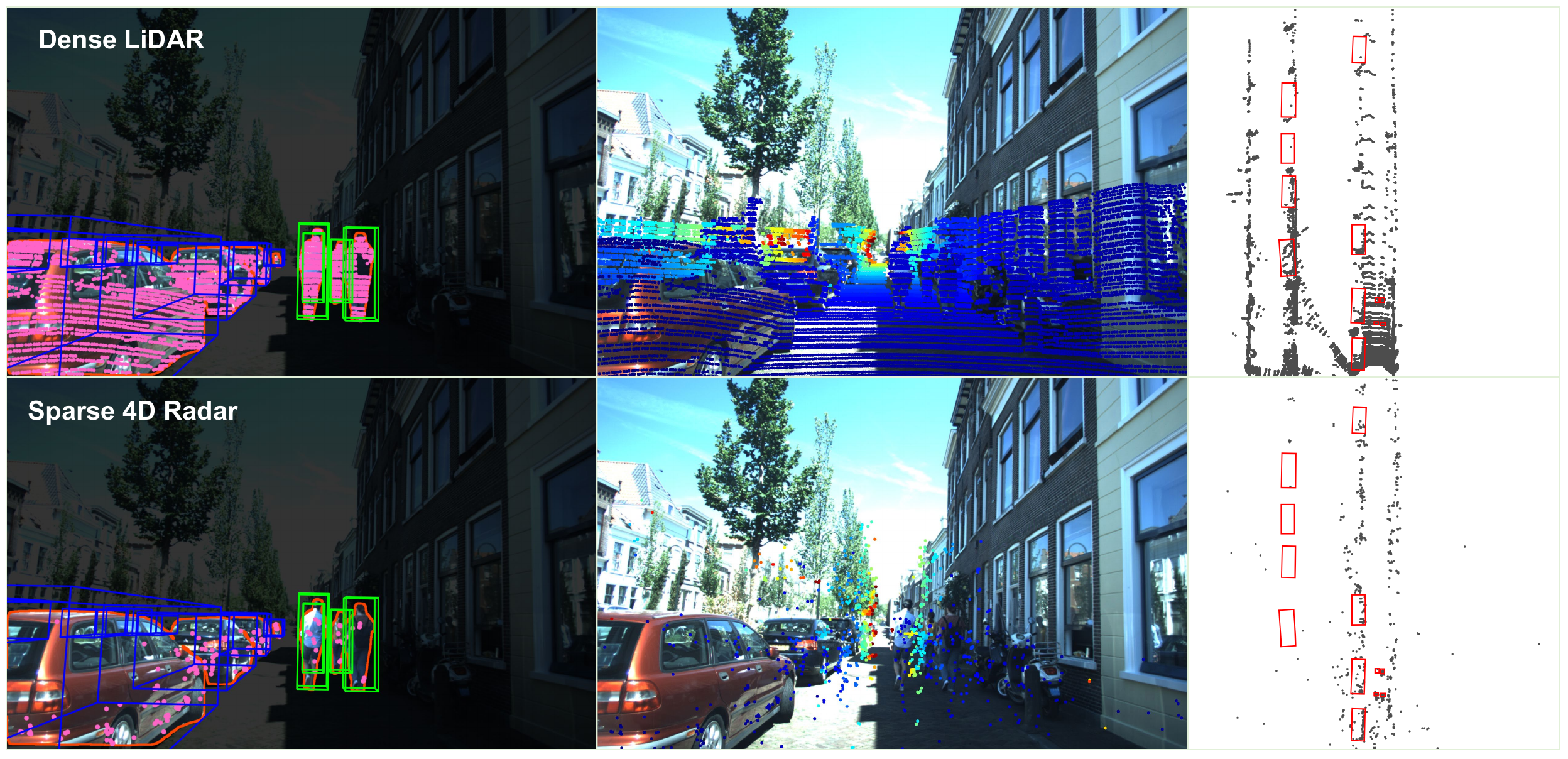}
    \caption{Visualization comparison of LiDAR and 4D radar on the VoD dataset: the first row shows LiDAR, and the second row shows 4D radar. The first column displays 3D ground truth boxes and the point cloud projection onto the foreground mask, with purple points indicating valid points providing object information. The second and third columns show the point cloud projection onto the perspective View and bird's-eye view, respectively. Dense LiDAR provides strong geometry, while sparse 4D radar only provides weak geometry.}
    \label{fig:LIDAR&RADAR}
\end{figure}

However, 4D radar data remain sparse and noisy due to limited resolution and multi-path effects \cite{sparse}, making standalone detection difficult and increasing the importance of fusion. Unlike LiDAR–camera fusion, which benefits from strong geometric priors, radar–camera fusion must address radar’s inherently weak geometry. Although LiDAR provides reliable instance cues crucial for 3D object detection, such strategies become ineffective under radar settings, where weak geometric signals make instance mining in BEV space challenging. As a result, methods like IS-Fusion \cite{IS-Fusion}, which rely on LiDAR’s strong geometry and extract instance features directly from BEV, tend to fail when BEV features become blurred after radar-based view transformation. This motivates the need for a mechanism that compensates for weak geometry using perspective-view information. To address this, we introduce a cross-view activation pipeline that injects 2D instance cues into BEV space to robustly activate instance-relevant regions despite sparse and noisy radar signals. Fig. \ref{fig:LIDAR&RADAR} shows the visualization comparison of LiDAR and 4D radar, where 4D radar provides only limited information compared to LiDAR.

While the above fundamental challenge highlights the necessity of stronger instance mining for radar–camera fusion, existing approaches in both paradigms still fall short. Recent advancements in radar-camera fusion have significantly boosted 3D object detection performance. These approaches can be roughly categorized into two paradigms: BEV-level fusion \cite{CRN, LXL, RCFusion, RCBEVDet, SGDet3D} and perspective-level fusion \cite{CRAFT, RADIANT, centerfusion}, each with its advantages and inherent limitations. On one hand, BEV-level fusion approaches (see Fig. {\ref{fig:comparison}(a)}) conduct view transformation to fill camera feature into the 3D volume, which are then fused with radar feature to generate unified scene feature in the bird's-eye (BEV) to enable global scene understanding. Despite effectiveness, these approaches suffer from weak instance focus. The equal treatment of foreground and background feature during the transformation process reduces feature contrast, causing objects to be overshadowed by background interference \cite{sparsefusion}, underscoring the necessity of enhancing instance-related feature within the scene. On the other hand, perspective-level fusion approaches (see Fig. {\ref{fig:comparison}(b)}) leverage 2D object detection to generate proposals \cite{cascade_maskrcnn}, which are refined by aligning radar feature to improve detection accuracy \cite{centerfusion}. However, their performance is limited by the cascaded network design inherent in proposal generation \cite{IS-Fusion, TMMFARPNet}. Furthermore, since the fusion occurs at the instance level, these approaches lack a holistic understanding of the scene \cite{RCMFusion}. Thus, neither BEV-level nor perspective-level approaches provide an effective mechanism to bridge scene understanding and robust instance activation, especially under weak radar geometry.

To mitigate the limitations introduced by the weak geometry of radar and boost instance awareness within the scene, we propose SIFormer (see Fig. \ref{fig:comparison}(c)), which effectively combines the complementary strengths of both perspective-level and BEV-level fusion paradigms by consistently enhancing instance awareness within the scene. Its core contribution lies in a cross-view activation and enhancement pipeline that (1) filters scene noise for feature initialization, (2) activates instance cues via perspective features, and (3) constructs a unified instance-aware BEV representation specifically suited for sparse radar. Specifically, we first employ sparse scene integration (SSI) to filter out interference feature via segmentation and depth-guided localization during view transformation, thereby focusing on regions of interest. Then, we introduce cross view correlation (CVC) to boosting instance awareness within the scene, enabling deep interactions between scene and instance feature by bridging bird's-eye view feature and perspective view feature. Finally, we design an instance enhance attention (IEA) module to aggregate image semantics and radar geometry from multi-modal feature, which ensuring robust perception. Experimental results on the TJ4DRadSet \cite{TJ4D} and View-of-Delft (VoD) \cite{VoD} datasets demonstrate that SIFormer effectively fuses 4D radar and camera data, outperforming existing approaches. In addition, experiments on the nuScenes \cite{nuscenes} dataset show that our model can also be adapted to 3D radar, achieving comparable performance. Our contributions can be summarized as follows:
\begin{itemize}
\item We propose SIFormer, a scene-instance aware transformer for 3D object detection. It is the first work to enhance instance awareness through cross-view correlation to mitigate the weak geometric consistency of radar. Extensive experiments demonstrate state-of-the-art performance.
\item We devise sparse scene integration (SSI) to filter out irrelevant features during view transformation, enabling the model to focus on regions of interest while preserving global scene understanding.
\item We design cross view correlation (CVC) to enhance instance awareness by bridging BEV and perspective-view features, enabling deep interaction between scene-level and instance-level representations. 
\item We introduce instance enhance attention (IEA) to effectively aggregate multi-modal semantic and geometric information, further reinforcing instance features and ensuring robust perception.
\end{itemize}

\section{Related Work}
\subsection{Single-Modality 3D Object Detection}
Camera and 4D radar are widely used in 3D object detection due to their affordability and complementary sensing capabilities. However, relying on a single modality brings fundamental limitations in both accuracy and robustness.

Camera-based 3D object detection approaches can be divided by their processing strategies. Approaches like BEVDet \cite{bevdet} and OFT \cite{OFT} lift image features into the BEV space but often suffer from depth ambiguity. BEVDepth \cite{BEVDepth} introduces LiDAR supervision to improve depth estimation, while transformer-based models such as BEVFormer \cite{BEVFormer} and DFA3D \cite{DFA3D} further enhance spatial reasoning. Simple-BEV \cite{SimpleBEV} additionally highlights challenges such as sparse features in distant regions and weak semantics in nearby areas. Other approaches, including FCOS3D \cite{fcos3d}, directly predict 3D bounding boxes from images without BEV construction, but often lack reliable depth cues. DETR3D \cite{detr3d} partially mitigates this by using sparse 3D queries with cross-attention mechanisms.

Radar-based approaches leverage 4D radar’s robustness in adverse conditions, but typically represent radar as sparse and noisy point clouds. Although approaches like RPFA-Net \cite{RPFANet}, SMURF \cite{SMURF} and LGDD \cite{LGDD} attempt to improve feature extraction, the inherent sparsity and weak geometric consistency remain significant challenges. As a result, radar-only detection still suffers from limited accuracy.

In summary, camera-based approaches lack accurate geometry, while radar-based approaches suffer from sparse and noisy measurements with limited semantic information. These limitations highlight radar-camera fusion as a promising direction for improving the reliability and accuracy of 3D perception. However, unlike LiDAR-camera fusion, which benefits from LiDAR’s strong geometric priors, radar provides only weak geometry, making effective fusion more challenging. To address this, SIFormer introduces cross-view correlation to enhance instance awareness within the scene, fully leveraging the complementary strengths of radar and camera.

\subsection{Multi-Modality 3D Object Detection}

\begin{figure}[ht]
    \centering
    \includegraphics[width=1.0\linewidth]{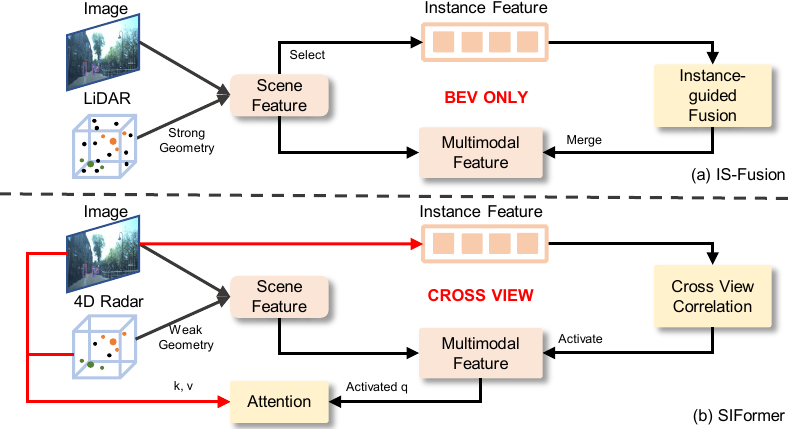}
    \caption{Comparison between IS-Fusion and SIFormer. (a) IS-Fusion mines instance features directly from scene features. (b) SIFormer employs cross-view correlation to improve radar-camera fusion, addressing weak radar geometry by activating instance awareness using 2D instance features.}
    \label{fig:ISFuison}
\end{figure}

Multi-modal fusion, particularly involving LiDAR or radar with cameras, has been widely explored to overcome the limitations of single-modality 3D object detection. Among them, LiDAR-camera fusion has received extensive attention due to the complementary nature of dense geometric information from LiDAR and rich semantic cues from cameras. Early works project image features onto LiDAR points or voxels for joint reasoning \cite{pointpainting, BEVFusion, SimpleBEV}, while recent approaches adopt more advanced cross-modal interaction modules \cite{IS-Fusion, unitr, TransFusion, DeepFusion}. For example, TransFusion \cite{TransFusion} employs a two-stream transformer to align features from LiDAR and images, and DeepInteraction \cite{DeepInteraction} further enhances this alignment through dense modality interaction. IS-Fusion \cite{IS-Fusion} leverages BEV representations to generate instance-aware features, benefiting significantly from the strong geometric priors of LiDAR. However, such approaches rely heavily on precise spatial information, which does not generalize well to radar due to its inherent sparsity, noise, and weak geometric consistency. In the absence of strong geometry, it becomes essential to focus on regions of interest during view transformation and to mitigate inaccurate BEV features caused by inherent limitations of radar  through a perspective-view bridging mechanism, as shown in Fig \ref{fig:ISFuison}.

Radar-camera fusion offers a more affordable and weather-resilient alternative, combining radar’s robustness and depth perception with camera’s high-resolution semantic information. Existing radar-camera fusion approaches can be broadly categorized into BEV-level and perspective-level approaches \cite{CRN, LXL, RCFusion, RCBEVDet, SGDet3D, CRAFT, RADIANT, centerfusion, RaGS, Wavelet, Doracamom}. BEV-level fusion aims to build a global BEV representation. For example, RCFusion \cite{RCFusion} applies OFT for feature lifting but overlooks uneven projection ray distribution; LXL \cite{LXL} incorporates radar occupancy to improve feature alignment; CRN \cite{CRN} combines LSS with radar data to enhance spatial reasoning. RCBEVDet \cite{RCBEVDet} uses a dual-stream RCS-aware design, and SGDet3D \cite{SGDet3D} employs cross-modal attention. However, these approaches often rely on depth prediction and view transformation, and typically ignore perspective-level features, resulting in feature blurring. Perspective-level fusion, on the other hand, focuses on accurate object localization. approaches like CenterFusion \cite{centerfusion} and CRAFT \cite{CRAFT} integrate radar and camera features at object centers, while RADIANT \cite{RADIANT} emphasizes explicit radar-camera association. FUTR3D \cite{FUTR3D} introduces sparse queries for multi-modal fusion. Despite their fine-grained accuracy, these approaches often lack global contextual understanding and suffer from cascaded designs that limit joint optimization \cite{IS-Fusion, TMMFARPNet, TMMDAM}. To address the limitations of both paradigms, we propose a unified radar-camera fusion model that bridges BEV-level and perspective-level representations.

\section{Method}
\subsection{Overview}
\begin{figure*}[ht]
    \centering
    \includegraphics[width=0.90\linewidth]{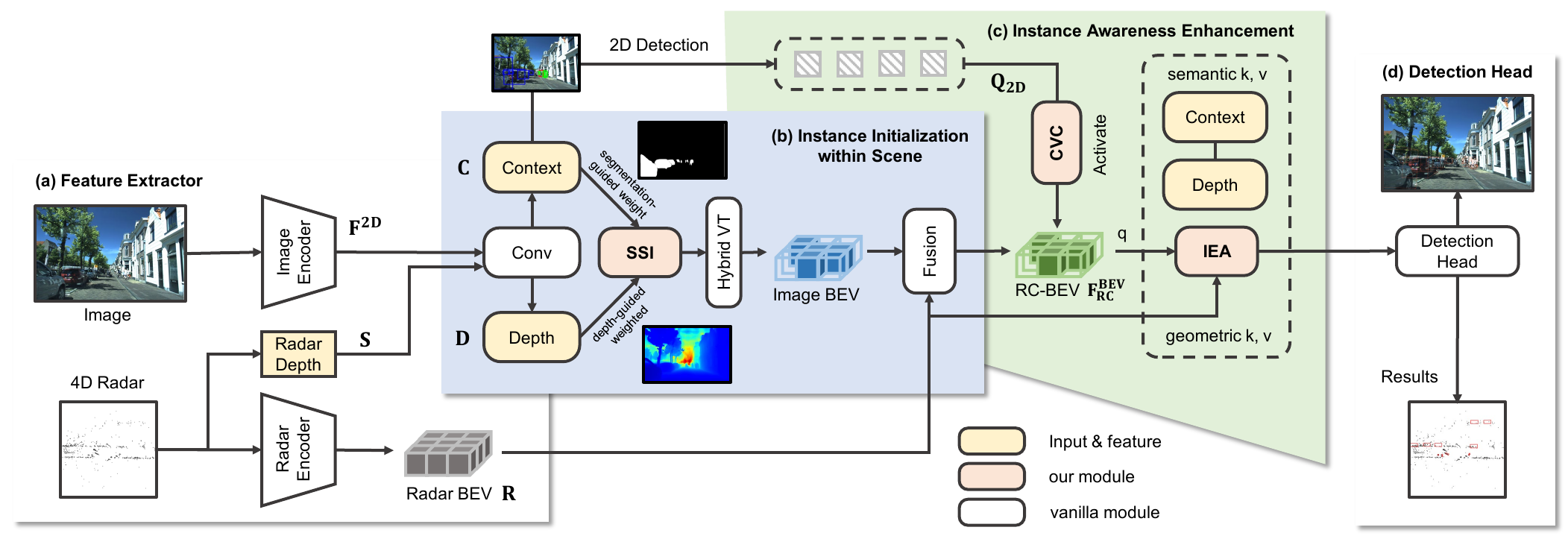}
    \caption{Architecture of our SIFormer. (a) The feature extractor extracts 4D radar and image feature from raw data. (b) The instance initialization stage filters out irrelevant features during view transformation via segmentation and depth-guided localization to focus on regions of interest introduces, while achieving global scene understanding. (c) The instance awareness enhancement stage leverages cross view correlation (CVC) to bridge perspective view instance feature with bird’s-eye view scene feature, followed by the instance enhance attention (IEA) module for further refinement, producing fused feature across scene and instance levels. (d) The decoder head for 3D object detection.}
    \label{fig:framework}
\end{figure*}

Fig. \ref{fig:framework} illustrates the architecture of our SIFormer that consists of four modules, including feature extractor, instance initialization within scene, instance awareness enhancement, and detection head. First, as in Section \ref{sec:Feature Extractor}, the feature extractor processes raw data to extract radar and image features. 

Then, we initialize instance feature within the scene based on BEV-level fusion, as in Section \ref{sec:BEV-level Fusion}. Specifically, we combines two view transformation techniques, improves view transformation accuracy by enabling accurate depth estimation and precisely filling camera feature into the 3D volume. During this process, sparse scene integration (SSI) filters background interference feature via perspective view segmentation and depth-guided localization to focus on regions of interest, which enables the model to focus on regions of interest while preserving global scene understanding. The resulting BEV feature map is further fused with the radar BEV feature map to produce feature RC-BEV.

Subsequently, perspective-level fusion is leveraged to further enhance instance awareness feature within the scene, which refers to Section \ref{sec:perspective-level Fusion}. The cross view correlation (CVC) connects instance feature with scene feature, enabling deep interactions with all potential activated regions of interest. This activated BEV representations then serves as improved queries in instance enhance attention (IEA) for further refinement, where each candidate instance aggregates image semantics and radar geometry for ensuring robustness.  

Finally, we feed the instance-boosted BEV-level feature into the detection head, based on \cite{PointPillars}, for 3D object detection, as in Section \ref{sec:Loss Function}. 

% SIFormer leverages perspective-level fusion to resolve the lack of instance awareness in BEV-level fusion and utilizes BEV-level fusion to overcome the limited global perception in perspective-level fusion. Notably, the novel cross view correlation (CVC) effectively boosts instance awareness within the scene for 3D object detection with 4D radar and camera. Consequently, our method effectively addresses the limitations of existing fusion paradigms while combining their strengths by consistently focus on instance within the scene.

\subsection{Feature Extractor}\label{sec:Feature Extractor}
The feature extractor extracts information from raw data. The image encoder consists of a backbone network ResNet50 to extract multi-scale feature and a feature pyramid network (FPN) to fuse them. We use $\mathbf{F}^{\text{2D}} \in \mathbb{R}^{H \times W \times C}$ to represent the extracted 2D image feature, where $C$ denotes the number of channel, and $(H, W)$ denotes the feature resolution. 

For radar data, we use the RadarPillarNet from \cite{RCFusion} to obtain the radar feature map in BEV domain. The preprocessed 4D radar points represented by 3D coordinates $x, y, z$, radar cross section $RCS$ or signal-to-noise ratio $SNR$, and absolute radial Doppler velocity $v_{rc}$ are encoded through 2D sparse convolution and passed through 2D convolution layers to obtain the radar BEV feature, denoted as $\mathbf{R} \in \mathbb{R}^{X \times Y \times C}$, where $(X, Y)$ denotes the BEV spatial resolution. Meanwhile, we project the 4D radar point cloud into the perspective view to form a sparse radar depth $\mathbf{S} \in \mathbb{R}^{X \times Y \times 1}$, where the last dimension represents the depth of the corresponding projection points on the image plane. In the remaining sections, we keep that the feature dimensions of the perspective view and bird's-eye view are consistent with $\mathbf{F}^{\text{2D}}$ and $\mathbf{R}$, respectively.

\subsection{Instance Initialization within Scene}\label{sec:BEV-level Fusion}
Existing BEV-level fusion approaches typically apply a single view transformation to project camera features into a 3D volume. However, this process often lacks instance-background separation and suffers from limited spatial accuracy. In contrast, we first combine semantic features from images with geometric cues from sparse radar depth to improve depth estimation and object localization. To address the limitations of prior works, such as sparse feature coverage in distant regions \cite{LSS} and under-utilization of nearby semantics \cite{SimpleBEV}, we propose a hybrid view transformation strategy. Furthermore, we introduce sparse scene integration (SSI), which leverages depth-guided localization to filter out background noise and geometric inaccuracies during view transformation, thereby enhancing instance awareness within the scene.

Specifically, the image feature $\mathbf{F}^{\text{2D}}$ is first processed through convolutional blocks to obtain a context feature map $\mathbf{C} \in \mathbb{R}^{H \times W \times C}$. Simultaneously, $\mathbf{F}^{\text{2D}}$ and sparse radar depth $\mathbf{S}$ are jointly fed into a convolutional fusion block to extract complementary cues from both modalities, yielding a discrete depth probability map $\mathbf{D} \in \mathbb{R}^{H \times W \times D}$, where $D$ denotes the number of predefined depth bins. Then, we perform hybrid view transformation strategy to construct 3D volume.

\begin{figure}[ht]
    \centering
    \includegraphics[width=\linewidth]{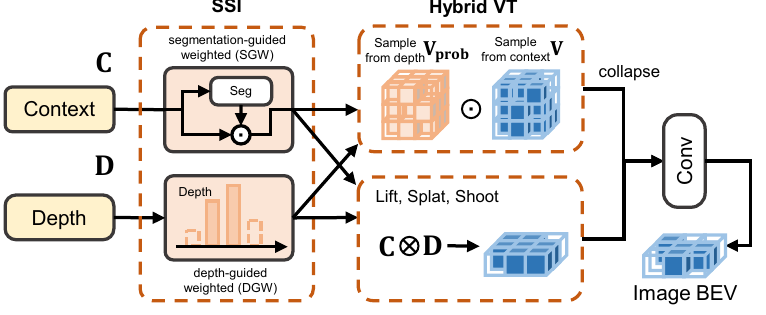}
    \caption{The detailed illustration of our instance initialization within scene stage. We employ sparse scene integration (SSI) to update depth and context, then fed then into hybrid view transformation to provide image BEV feature.}
    \label{fig:viewtransform}
\end{figure}

Following \cite{LSS}, we compute the outer product of $\mathbf{D}$ and $\mathbf{C}$ and apply voxel pooling to generate an initial BEV feature map, as shown in Fig. \ref{fig:viewtransform}. In parallel, inspired by Simple-BEV \cite{SimpleBEV}, we construct a set of predefined voxels with shape $X \times Y \times Z$ in the radar coordinate system and define virtual points \cite{SimpleBEV} at the center of each voxel. These virtual points are projected into the image feature map, from which we bilinearly sample corresponding features to form a 3D feature volume $\mathbf{V} \in \mathbb{R}^{X \times Y \times Z \times C}$. Unlike prior works, we also project the virtual points into the depth probability map $\mathbf{D}$ to retrieve the depth distribution along each projection ray. Using the depth values of the virtual points, we perform linear interpolation on this distribution to obtain a 3D probability volume $\mathbf{V}_{\text{prob}} \in \mathbb{R}^{X \times Y \times Z}$. The final volume is computed by taking the dot product between $\mathbf{V}$ and $\mathbf{V}_{\text{prob}}$, followed by a multi-layer perceptron (MLP) to reduce the height dimension and generate a collapsed BEV feature.

Finally, we apply convolutional layers to fuse the two BEV features and obtain the image BEV representation. This feature is further fused with the radar BEV feature $\mathbf{R}$ using a cross-modality fusion block, resulting in the final radar-camera BEV feature $\mathbf{F}_{\text{RC}}^{\text{BEV}}$.

\textbf{Sparse Scene Integration (SSI).} Due to the geometry inaccuracy and background semantic feature interference, we devise to filter out interference feature during view transformation and increase instance awareness both semantically and geometrically, thereby enhancing the feature contrast in the scene feature. To achieve this, we update context $\mathbf{C}$ and depth $\mathbf{D}$ during the process of view transformation. Specifically, for semantics feature filtering, we use a lightweight segmentation network to predict the foreground regions in the perspective view. The predicted mask is then used to re-weight the context feature,w which we refer to as segmentation-guided weighted (SGW). For geometry feature filtering, we use depth-guided weighted (DGW) to retain the top-K discrete depth probabilities and discard the rest, which helps mitigate the impact of imprecise depth estimation and prevents camera feature from being filled into low-probability regions, thereby reducing interference. The update process can be expressed as
\begin{equation}
\mathbf{C} \leftarrow \mathbf{C} \odot \mathbf{M},
\end{equation}
\begin{equation}
\mathbf{D} \leftarrow \mathtt{Normalize}(\mathbf{D}_\text{K}),
\end{equation}
where $\mathbf{M} \in \mathbb{R}^{H \times W}$ is the predicted mask, $\mathbf{D}_\text{K}$ is the top-K values in $\mathbf{D}$, and $\mathtt{Normalize}$ denotes min-max normalization. In this work, we simply retain the top 25\% of the depth bins.

\subsection{Instance Awareness Enhancement} \label{sec:perspective-level Fusion}
\begin{figure*}[ht]
    \centering
    \includegraphics[width=0.90\linewidth]{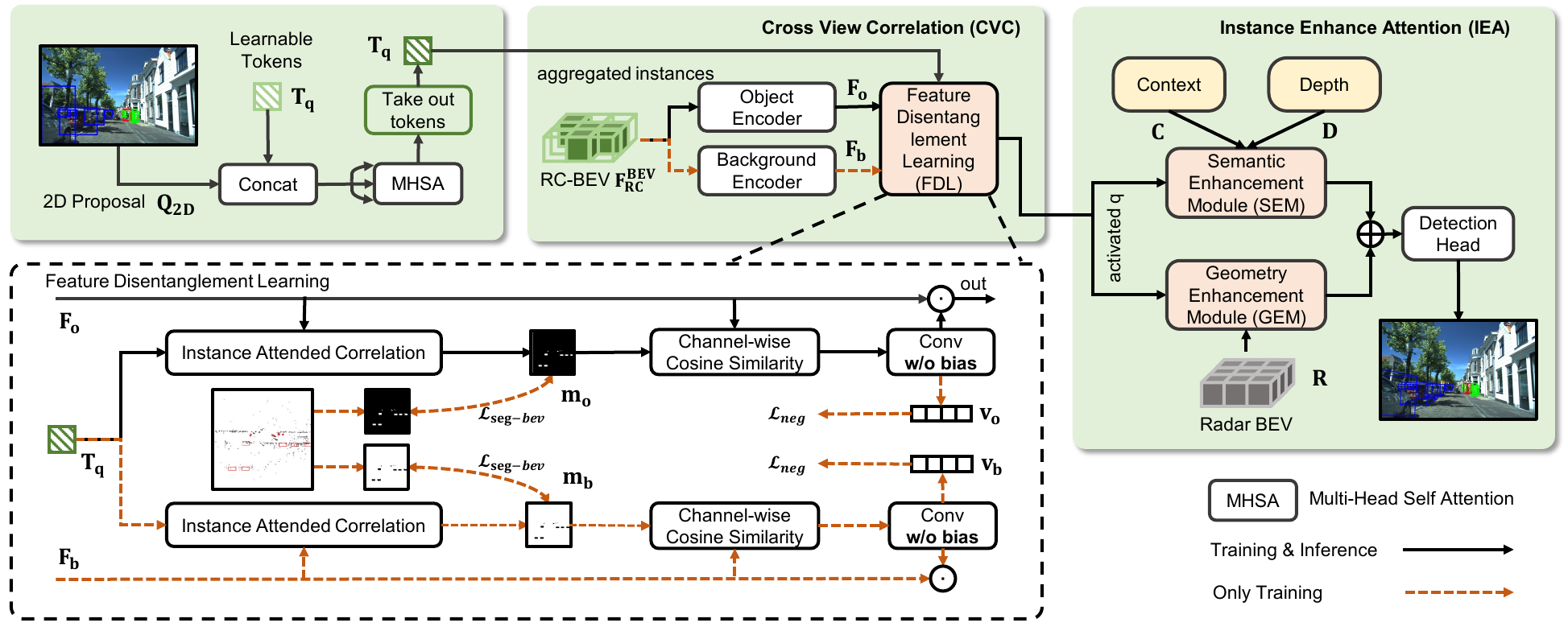}
    \caption{The illustration of our instance awareness enhancement stage. We first employ cross view correlation (CVC) to activate all potential regions of interest within scene feature using a learnable token. To be specific, the instance attended correlation connects the aggregated instances with the RC-BEV through matrix operations, producing a correlation map, which is then used to calculate the cosine similarity with the scene feature. Consequently, the output of CVC serve as improved queries for the subsequent instance enhance attention (IEA), facilitating further aggregation of semantics and geometry. }
    \label{fig:instancelevel}
\end{figure*}

Previous perspective-level fusion approaches struggle to integrate global BEV-level feature \cite{IS-Fusion, TMMFARPNet}. Although \cite{centerfusion} uses frustum-based techniques to filter relevant feature and enhance interactions, they still lack comprehensive scene perception. To address this, we devise cross view correlation (CVC) to enable deep scene-instance interaction with a learnable token, thereby facilitating global perception in perspective-level fusion while building upon BEV-level fusion. Then, instance enhance attention (IEA) further exploits image semantics and radar geometry for each candidate instance within the enhanced scene feature, producing fused feature across both levels. This process overcomes the limitations of \cite{CRAFT}, which also utilizes transformers for feature aggregation but is constrained by cascaded network design. As a result, our CVC and IEA effectively address the lack of instance awareness in BEV-level fusion and the limited global perception in perspective-level fusion, as shown in Fig. \ref{fig:instancelevel}.

The gap between the perspective view and bird’s-eye view has long existed, with limited research addressing their connection. To the best of our knowledge, our CVC is the first module to connect these two views, significantly enhancing the interaction between instance feature in both perspectives. Moreover, as improved feature queries, it strengthens the feature aggregation effect of IEA, with both components complementing each other and forming an inseparable relationship.

\begin{figure}[ht]
    \centering
    \includegraphics[width=\linewidth]{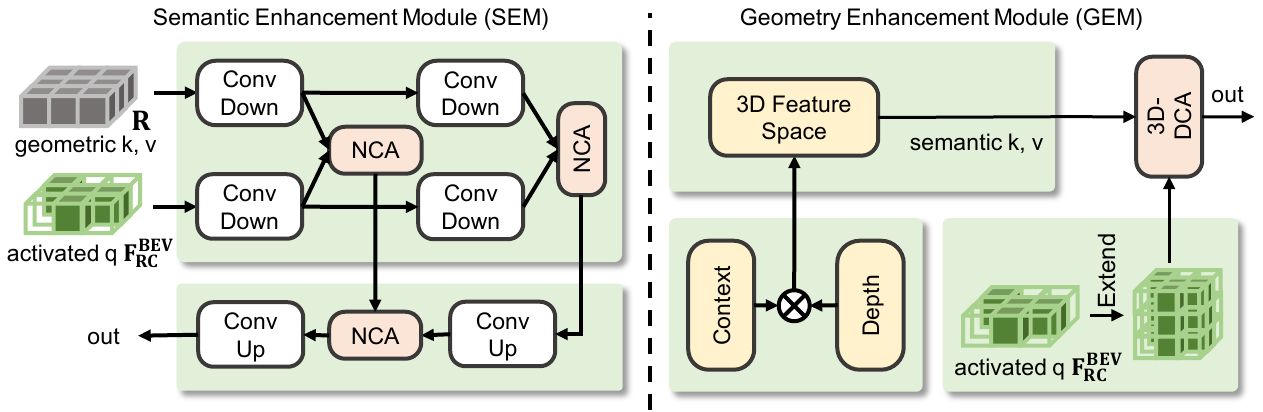}
    \caption{The illustration of our instance enhancement attention (IEA), consisted of semantic enhancement module and geometry enhancement module.}
    \label{fig:attention}
\end{figure}

\textbf{Cross View Correlation (CVC).}
Given the success of 2D object detection and its relatively higher recall compared to 3D object detection \cite{far3d}, we aim to extend it by enabling scene-instance interaction between perspective view instance feature and BEV scene feature. Compared with LiDAR-camera fusion model IS-Fusion \cite{IS-Fusion}, SIFormer enables more effective instance generation by cross view mapping as radar data lacks the strong geometric priors that LiDAR provides. By using cross-view correlation, SIFormer overcomes the feature blurring challenges in BEV-only approaches by activating instance-related region. 

Initially, we use Cascade Mask R-CNN \cite{cascade_maskrcnn} to detect 2D instances from the context $\mathbf{C}$ as proposals. A learnable token, $\mathbf{T}_{\text{q}} \in \mathbb{R}^{1 \times C}$, is then concatenated with these proposals to extract perspective-level information through self-attention. Simultaneously, the RC-BEV feature map, $\mathbf{F}_{\text{RC}}^{\text{BEV}}$, is passed through object and background encoders to extract object-specific and background-specific information, respectively. Next, through feature disentanglement learning (FDL), $\mathbf{T}_{\text{q}}$ effectively transfers local perspective-level information to the global scene level. Specifically, in the instance-attended correlation, $\mathbf{T}_{\text{q}}$ interacts with the object and background BEV feature to generate object and background segmentation masks. These masks are then used to compute channel-wise cosine similarity with the BEV feature, verifying whether the activated regions of $\mathbf{T}_{\text{q}}$ align with the ground-truth object and background occupied regions. By leveraging 2D object detection and the learnable token, our CVC bridges cross views and enable the activation of all potential instance-related regions within the scene feature. This process ultimately generates instance-enhanced BEV feature, denoted as  $\mathbf{F}_{\text{Activated}}^{\text{BEV}}$.

Taking the object BEV feature map $\mathbf{F}_{\text{o}}$ as an example, the instance attended correlation connects perspective-level feature with BEV-level feature via matrix operations, which produces a correlation map
$\widehat{\mathbf{m}}_{\mathrm{o}} \in \mathbb{R}^{X \times Y}$,
\begin{equation}
\widehat{\mathbf{m}}_{\mathrm{o}}=\sigma(\mathbf{F}_{\text{o}}\mathbf{T}_{\mathrm{q}}^{\top}),
\end{equation}
where $\sigma(\cdot)$ denotes the sigmoid function. We then utilize ground-truth object occupied mask ${\mathbf{m}}_{\mathrm{o}}$ to supervise the generation of $\widehat{\mathbf{m}}_{\mathrm{o}}$. The supervision aims to maximize the expression of object feature on the BEV correlation map, thus further leading $\mathbf{T}_{\text{q}}$ to extract perspective-level information. Then we calculate the cosine similarity between $\widehat{\mathbf{m}}_{\mathrm{o}}$ and $\mathbf{F}_{\text{o}}$ across channels to create a similarity vector ${\mathbf{v}}_{\mathrm{o}} \in \mathbb{R}^{C}$. Each element in the vector represents the similarity between a channel feature map and the $\widehat{\mathbf{m}}_{\mathrm{o}}$. Since $\widehat{\mathbf{m}}_{\mathrm{o}}$ is expected to highlight the object regions, thus we enforce each element in this vector to approximate 1.0, encouraging the BEV feature map to shape itself towards the ground-truth occupied regions. Finally, we perform dot product between $\mathbf{F}_{\text{o}}$ and ${\mathbf{v}}_{\mathrm{o}}$ to obtain output. We further construct a background counterpart as a part of cross view correlation to enhance the guidance. During the inference, the process can be expressed as
\begin{equation}
\mathbf{F}_{\text{Activated}}^{\text{BEV}} = \mathtt{CVC}(\mathbf{Q}_{\text{2D}}, \mathbf{F}_{\text{RC}}^{\text{BEV}}, \mathbf{T}_{\text{q}}),
\end{equation}
where $\mathtt{CVC}$ denotes the whole cross view correlation, $\mathbf{Q}_{\text{2D}}$ are the 2D proposal, i.e. instance feature. The corresponding background BEV feature map, correlation map and similarity vector are denoted as $\mathbf{F}_{\text{b}}$, $\widehat{\mathbf{m}}_{\mathrm{b}}$ and  $\widehat{\mathbf{v}}_{\mathrm{b}}$.

\begin{figure*}[t]
    \centering
    \includegraphics[width=\linewidth]{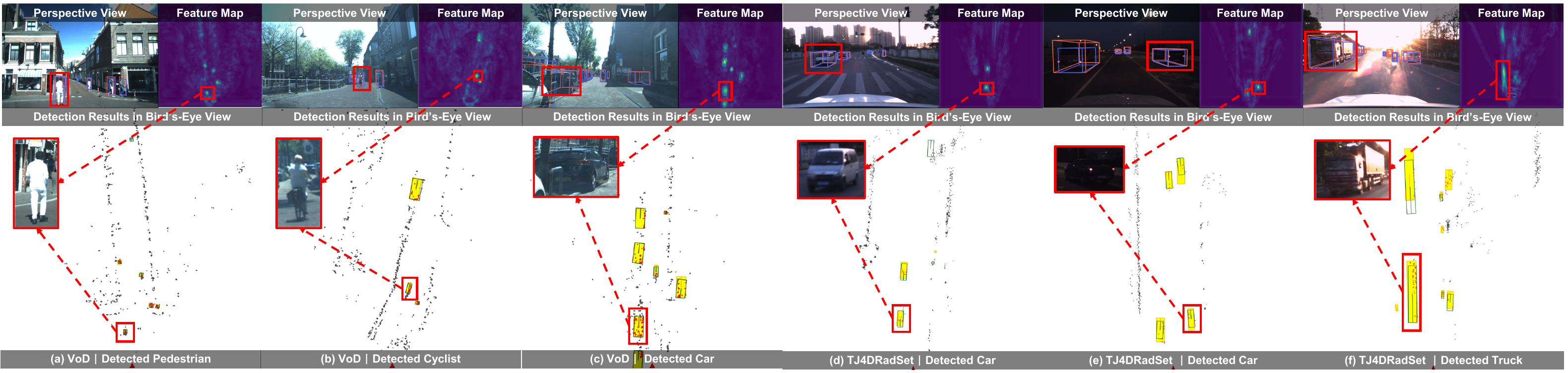}
    \caption{Visualization results on the VoD validation set ((a),(b),(c)) and TJ4DRadSet test set ((d),(e),(f)). Each figure corresponds to a frame. Orange and yellow boxes represent ground-truths in the perspective and bird's-eye view, respectively. Green and blue boxes indicate predicted results.}
    \label{fig:Visualization}
\end{figure*}

\begin{figure*}[t]
    \centering
    \includegraphics[width=\linewidth]{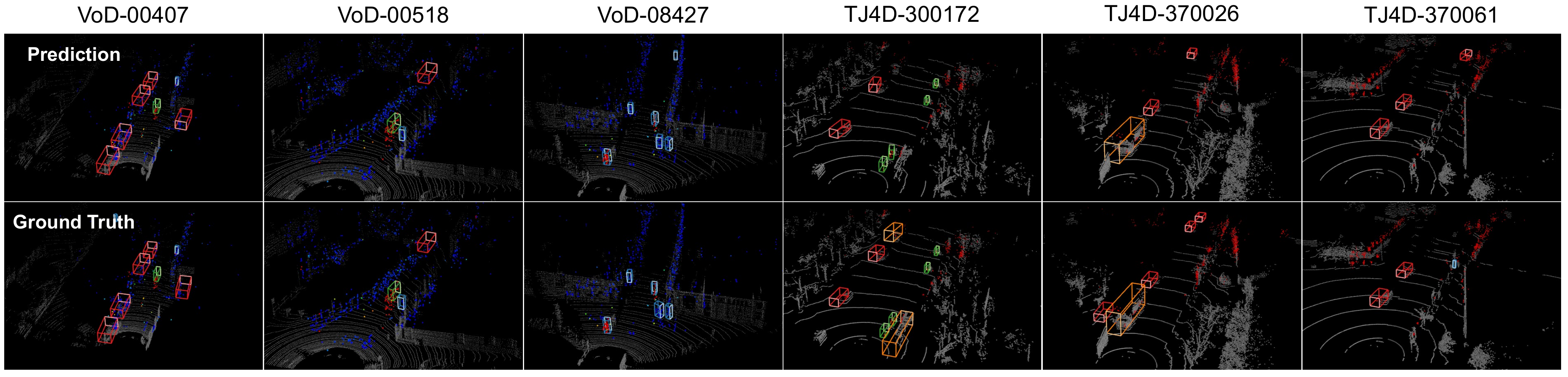}
    \caption{Qualitative presentation of SIFormer. The first row is prediction and the second row denotes the ground truth. White dots represent LiDAR, while colored dots represent 4D radar, with color differences indicating velocity variations. The comparison highlights the effectiveness of our SIFormer.}
    \label{fig:3Dvisualization}
\end{figure*}

\textbf{Instance Enhance Attention (IEA).} After our CVC, all potential regions of interest related to instance feature within the scene feature are effectively activated. Therefore, consistent with the concepts of forward and backward fusion in \cite{FBBEV} and \cite{DFA3D}, we believe that the enhanced BEV feature can serve as improved queries, enabling improved performance in the subsequent transformer-based interaction between image and radar feature. This approach differs from IS-Fusion\cite{IS-Fusion}, which only mines instance features from the bird’s-eye view and enhances itself in isolation, providing better robustness.Specifically, we use instance enhance attention (IEA) to further enhance BEV feature from both the semantic and geometric perspectives.

On one hand, we enhance the semantic information of the BEV feature using a semantic enhancement module. Following the process in \cite{DFA3D}, we first create the 3D feature space using context and depth, then extend the BEV queries in the pillar dimension. These queries are subsequently projected into the 3D feature space to perform 3D deformable cross-attention (3D-DCA), ultimately enabling more effective aggregation of the semantic information from the image data. On the other hand, we devise a geometry enhancement module that utilizes the rich radar occupancy information to assist in feature enhancement. For the input $\mathbf{F}_{\text{Activated}}^{\text{BEV}}$ and the previous radar feature $\mathbf{R}$, we employ a multi-scale fusion mechanism that combines the U-Net architecture and neighborhood cross attention (NCA) \cite{neighborhood} for efficient fusion, enabling the effective aggregation of geometric information from radar data.

As a result, with $\mathbf{F}_{\text{Activated}}^{\text{BEV}}$ already activating instance-related regions, we effectively aggregate image semantics and radar geometry for each candidate query through a transformer. The enhanced outputs are then summed, producing the final Instance-BEV $\mathbf{F}_{\text{Final}}^{\text{BEV}}$ that can be expressed as
\begin{equation}
\mathbf{F}_{\text{Final}}^{\text{BEV}} = \mathtt{SEM}(\mathbf{F}_{\text{Activated}}^{\text{BEV}}, \mathbf{D}, \mathbf{C}) + \mathtt{GEM}(\mathbf{F}_{\text{Activated}}^{\text{BEV}}, \mathbf{R}),
\end{equation}
where $\mathtt{SEM}$ denotes the semantic enhancement module and $\mathtt{GEM}$ denotes the geometry enhancement module.

\subsection{Loss Function} \label{sec:Loss Function}
We divide the loss into three parts: initialization stage loss, enhancement stage loss, and detection loss. For detection, we simply adopt $\mathcal{L}_{\text{det2D}}$ from \cite{cascade_maskrcnn} for 2D object detection and $\mathcal{L}_{\text{det3D}}$ from \cite{RCFusion} for 3D object detection.

The initialization stage loss loss includes perspective view foreground segmentation and discrete depth probability estimation. For segmentation, the processed results from Detectron2 combined with ground-truth 2D bounding boxes serve as ground-truth, and the loss $\mathcal{L}_{\text{seg-per}}$ from \cite{FBBEV} is used,  which incorporates binary cross-entropy and Dice loss. For depth estimation, LiDAR depth serves as ground truth, with the depth loss $\mathcal{L}_{\text{depth}}$ from \cite{BEVDepth}. The enhancement stage loss includes supervision for the object/background correlation map and similarity vector ($\mathbf{v}$). we use the ground-truth 3D bounding boxes to form the occupancy map as the ground truth to supervise correlation map using the segmentation loss $\mathcal{L}_{\text{seg-bev}}$ from \cite{FBBEV}. For the similarity vector $\mathbf{v}$, we use $\mathcal{L}_{\text{neg}}$ to encourage the BEV feature map to shape itself towards the ground-truth occupied region, which can be formulated as
\begin{equation}
\mathcal{L}_{\text{neg}}(\mathbf{v})=-\frac{1}{C}\sum_{i=1}^C \log(\sigma(\mathbf{v})),
\end{equation}
where $\sigma(\cdot)$ denotes the sigmoid function, and $C$ is the number of channel dimensions.

The total loss is formulated as
\begin{equation}
\begin{split}
\mathcal{L}_{\text{total}} = & \mathcal{L}_{\text{det3D}} + \mathcal{L}_{\text{det2D}} + \lambda_1 \mathcal{L}_{\text{depth}} \\
& + \lambda_2 \mathcal{L}_{\text{neg}} +  \lambda_3 (\mathcal{L}_{\text{seg-per}} + \mathcal{L}_{\text{seg-bev}}),
\end{split}
\end{equation}
where the hyperparameters $\lambda_1$, $\lambda_2$ and $\lambda_3$ balance the each loss term, respectively. In this work, we set all hyperparameters to 1.0 for simplicity.

\begin{table*}[ht]
    \belowrulesep=0pt
    \aboverulesep=0pt
    \centering
    \footnotesize
    \setlength{\tabcolsep}{4.8pt} 
    \caption{Comparison with state-of-the-art approaches on the validation set of VoD \cite{VoD}. In the Modality column, R denotes 4D radar and C denotes camera. The results of approaches marked with $\dagger$ indicate the use of extra LiDAR data. The best values are in bold, and the second-best values are underlined.}
	\renewcommand\arraystretch{1.1} %增加表格行距
	% \resizebox{1.0\linewidth}{!}{
    \begin{tabular}{c|c|cccc|cccc|c}
        \toprule[1.0pt]
        \multirow{2}{*}[-0.7ex]{\centering Approach} & \makebox[0.75cm]{\multirow{2}{*}[-0.7ex]{\centering Input}} & \multicolumn{4}{c|}{Entire Annotated Area (\%)} & \multicolumn{4}{c|}{Driving Corridor (\%)} & \multirow{2}{*}[-0.7ex]{\centering FPS} \\ 
        \cline{3-6} \cline{7-10}
         &  & \makebox[0.8cm]{Car} & \makebox[0.8cm]{Ped} & \makebox[0.8cm]{Cyc} & \makebox[0.8cm]{mAP} & \makebox[0.8cm]{Car} & \makebox[0.8cm]{Ped} & \makebox[0.8cm]{Cyc} & \makebox[0.8cm]{mAP} \\ \midrule
        ImVoxelNet (WACV 2022) \cite{imvoxelnet} & C & 19.35 & 5.62 & 17.53 & 14.17 & 49.52 & 9.68 & 28.97 & 29.39 & 11.1\\
        \midrule
        % PillarNeXt (CVPR 2023) \cite{pillarnext} & R & 30.81 & 33.11 & 62.78 & 42.23 & 66.72 & 39.03 & 85.08 & 63.61 & -\\
        % PointPillars (CVPR 2019) \cite{PointPillars} & R & 37.06 & 35.04 & 63.44 & 45.18 & 70.15 & 47.22 & 85.07 & 67.48 & 113.9 \\
        % CenterPoint (CVPR 2021) \cite{centerpoint} & R & 35.84 & 41.03 & 67.11 & 47.99 & 70.65 & 50.14 & 85.67 & 68.82 & 38.3\\
        % VoxelNeXt (CVPR 2023) \cite{VoxelNeXt} & R & 36.98 & 42.37 & 68.15 & 49.17 & 70.95 & 51.85 & 87.33 & 70.04 & 31.6\\
        RadarPillarNet (IEEE TIM 2023) \cite{RCFusion} & R & 39.30 & 35.10 & 63.63 & 46.01 & 71.65 & 42.80 & 83.14 & 65.86 & 98.8 \\
        \text{LXL - R} \text{(IEEE TIV 2023)} & R & 32.75 & 39.65 & 68.13 & 46.84 & 70.26 & 47.34 & 87.93 & 68.51 & 44.7\\
        SMURF (IEEE TIV 2023) \cite{SMURF} & R & 42.31 & 39.09 & 71.50 & 50.97 & 71.74 & 50.54 & 86.87 & 69.72 & -\\
        \midrule
        FUTR3D (CVPR 2023) \cite{FUTR3D} & R+C & 46.01 & 35.11 & 65.98 & 49.03 & 78.66 & 43.10 & 86.19 & 69.32 & 7.3 \\
        BEVFusion (ICRA 2023) \cite{BEVFusion} & R+C & 37.85 & 40.96 & 68.95 & 49.25 & 70.21 & 45.86 & 89.48 & 68.52 & 7.1\\
        RCFusion (IEEE TIM 2023) \cite{RCFusion} & R+C & 41.70 & 38.95 & 68.31 & 49.65 & 71.87 & 47.50 & 88.33 & 69.23 & 9.0\\
        % TL-4DRCF (IEEE Sensors 2024) \cite{TL-4DRCF} & R+C & 43.71 & 40.11 & 64.22 & 49.35 & 79.49 & 53.76 & 76.50 & 69.92 & -\\
        RCBEVDet (CVPR 2024) \cite{RCBEVDet} & R+C & 40.63 & 38.86 & 70.48 & 49.99 & 72.48 & 49.89 & 87.01 & 69.80 & -\\
        IS-Fusion (CVPR 2024) \cite{IS-Fusion} & R+C & 48.57 & 46.17 & 68.48 & 54.40 & 80.42 & 55.50 & 88.33 & 74.75 & 15.1\\
        % UniBEVFusion (arXiv 2024) \cite{UniBEVFusion} & R+C & 42.22 & 47.11 & 72.94 & 54.09 & 72.10 & 57.71 & 93.29 & 74.37 & -\\
        LXL (IEEE TIV 2023) \cite{LXL} & R+C & 42.33 & 49.48 & \textbf{77.12} & 56.31 & 72.18 & 58.30 & 88.31 & 72.93 & 6.1\\ 
        ${\text{SGDet3D}^{\dagger}}$ (IEEE RAL 2024) \cite{SGDet3D} & R+C & 53.16 & 49.98 & 76.11 & 59.75 & 81.13 & 60.91 & \textbf{90.22} & \underline{77.42} & 9.2\\
        % HyDRa (ICRA 2025) \cite{HyDRa} & R+C & 52.83 & \textbf{56.57} & 73.25 & \underline{60.88} & 80.65 & \underline{62.90} & 87.43 & 76.99 & - \\
        % MSSF-PP (arXiv 2024) \cite{MSSF} & R+C & 60.96 & 51.28 & \textbf{77.69} & 63.31 & 90.60 & 60.39 & 88.35 & 79.78 & 13.9\\
        \rowcolor{gray!20} SIFormer (\textbf{Ours}) & R+C & \underline{53.75} & \underline{50.68} & 76.12 & \underline{60.18} & \underline{81.16} & 60.50 & \underline{90.15} & 77.27 & 6.9 \\ 
        % \rowcolor{gray!20} PointPillars (CVPR 2019) \cite{PointPillars} & L & 60.48 & 45.78 & 69.16 & 58.47 & 90.27 & 61.81 & 87.13 & 79.74 & 63.8 \\
        \rowcolor{gray!20} ${\text{SIFormer}^{\dagger}}$ (\textbf{Ours}) & R+C & \textbf{61.49} & \textbf{51.39} & \underline{77.09} & \textbf{63.32} & \textbf{90.67} & \textbf{68.60} & 89.92 & \textbf{83.06} & 6.9 \\ 
        \bottomrule[1.0pt]
    \end{tabular}
    \label{tab:VoD}
\end{table*}

\begin{table*}[ht]
    \belowrulesep=0pt
    \aboverulesep=0pt
    \centering
    \footnotesize
    \setlength{\tabcolsep}{5pt}
    \caption{Comparison with state-of-the-art approaches on the test set of TJ4DRadSet \cite{TJ4D}. The other settings are the same as in Table \ref{tab:VoD}.}
        \renewcommand\arraystretch{1.1} %增加表格行距
	% \resizebox{1.0\linewidth}{!}{
    \begin{tabular}{c|c|ccccc|ccccc}
        \toprule[1.0pt]
        \multirow{2}{*}[-0.7ex]{\centering Approach} & \makebox[0.75cm]{\multirow{2}{*}[-0.7ex]{\centering Input}} & \multicolumn{5}{c|}{$\text{AP}_\text{3D}$ (\%)} & \multicolumn{5}{c}{$\text{AP}_\text{BEV}$ (\%)} \\
        \cline{3-7} \cline{8-12}
         &  & \makebox[0.5cm]{Car} & \makebox[0.5cm]{Ped} & \makebox[0.5cm]{Cyc} & \makebox[0.5cm]{Truck} & \makebox[0.5cm]{mAP} & \makebox[0.5cm]{Car} & \makebox[0.5cm]{Ped} & \makebox[0.5cm]{Cyc} & \makebox[0.5cm]{Truck} & \makebox[0.5cm]{mAP} \\ \midrule
        ImVoxelNet (WACV 2022) \cite{imvoxelnet} & C & 22.55 & 13.73 & 9.67 & 13.87 & 14.96 & 26.10 & 14.21 & 10.99 & 17.18 & 17.12 \\
        \midrule
        RadarPillarNet (IEEE TIM 2023) \cite{RCFusion} & R & 28.45  & 26.24 & 51.57 & 15.20 & 30.37 & 45.72 & 29.19 & 56.89 & 25.17 & 39.24 \\
        LXL - R (TIV 2023) \cite{LXL} & R & - & - & - & - & 30.79 & - & - & - & - & 38.42 \\
        SMURF (IEEE TIV 2023) \cite{SMURF} & R & 28.47 & 26.22 & \underline{54.61} & 22.64 & 32.99 & 43.13 & 29.19 & \textbf{58.81} & 32.80 & 40.98 \\
        \midrule
        FUTR3D (CVPR 2023) \cite{FUTR3D} & R+C & - & - & - & - & 32.42 & - & - & - & - & 37.51 \\
        BEVFusion (ICRA 2023) \cite{BEVFusion} & R+C & - & - & - & - & 32.71 & - & - & - & - & 41.12 \\
        LXL (IEEE TIV 2023) \cite{LXL} & R+C & - & - & - & - & 36.32 & - & - & - & - & 41.20 \\ 
        % IS-Fusion (CVPR 2024) \cite{IS-Fusion} & R+C & - & - & - & - & - & - & - & - & - & - \\
        RCFusion (IEEE TIM 2023) \cite{RCFusion} & R+C & 29.72 & 27.17 & \textbf{54.93} & 23.56 & 33.85 & 40.89 & 30.95 & \underline{58.30} & 28.92 & 39.76 \\
        UniBEVFusion (arXiv 2024) \cite{UniBEVFusion} & R+C & 44.26 & \underline{27.92} & 51.11 & 27.75 & 37.76 & 50.43 & 29.57 & 56.48 & 35.22 & 42.92 \\
        ${\text{SGDet3D}^{\dagger}}$ (IEEE RAL 2024) \cite{SGDet3D} & R+C & 59.43 & 26.57 & 51.30 & 30.00 & 41.82 & 66.38 & 29.18 & 53.72 & 39.36 & 47.16 \\
        \rowcolor{gray!20} SIFormer (\textbf{Ours}) & R+C & \underline{61.12} & 27.18 & 46.11 & \textbf{38.22} & \underline{43.15} & \underline{68.30} & \underline{31.47} & 46.46 & \textbf{45.64} & \underline{47.96} \\ 
        \rowcolor{gray!20} ${\text{SIFormer}^{\dagger}}$ (\textbf{Ours}) & R+C & \textbf{65.68} & \textbf{29.77} & 54.47 & \underline{36.12} & \textbf{46.51} & \textbf{71.59} & \textbf{36.06} & 55.15 & \underline{43.31} & \textbf{50.42} \\ 
        \bottomrule[1.0pt]
    \end{tabular}
    \label{tab:TJ4D}
\end{table*}

\begin{table}[ht]
    \belowrulesep=0pt
    \aboverulesep=0pt
    \centering
    \footnotesize
    \caption{Comparison of 3D object detection results on the nuScenes val set. L, C, and R represent LiDAR, Camera, and 3D Radar, respectively.}
    \renewcommand\arraystretch{1.2}
    \begin{tabular}{c|c|c|c|cc}
        \toprule[1.0pt] 
        \makebox[1.50cm]{Approach} & \makebox[0.50cm]{Input} & \makebox[1.00cm]{Backbone} & \makebox[1.00cm]{Image Size} & \makebox[0.50cm]{NDS$\uparrow$} & \makebox[0.50cm]{mAP$\uparrow$} \\
        \midrule
        BEVDet \cite{bevdet} & C & R50 & 256 × 704 & 39.2 & 31.2 \\
        BEVDepth \cite{BEVDepth} & C & R50 & 256 × 704 & 47.5 & 35.1 \\
        FB-BEV \cite{FBBEV} & C & R50 & 256 × 704 & 49.8 & 37.8 \\
        \midrule
        CenterFusion \cite{centerfusion} & R+C & DLA34 & 448 × 800 & 45.3 & 33.2 \\
        CRAFT \cite{CRAFT} & R+C & DLA34 & 448 × 800 & 51.7 & 41.1 \\
        RCBEVDet \cite{RCBEVDet} & R+C & DLA34 & 448 × 800 & 56.3 & 45.3 \\
        X3KD \cite{x3kd} & R+C & R50 & 256 × 704 & 53.8 & 42.3 \\
        CRN \cite{CRN} & R+C & R50 & 256 × 704 & 56.0 & \textbf{49.0} \\
        RCBEVDet \cite{RCBEVDet} & R+C & R50 & 256 × 704 & \underline{56.8} & 45.3 \\
        \rowcolor{gray!20} ${\text{SIFormer}^{\dagger}}$ (\textbf{Ours}) & R+C & R50 & 256 × 704 & \textbf{56.8} & \underline{46.0} \\
        \bottomrule[1.0pt]
    \end{tabular}
    \label{tab:nuscenes}
\end{table}

\begin{table}[t]
    \belowrulesep=0pt
    \aboverulesep=0pt
    \centering
    \footnotesize
    \caption{Ablation study on each component of SIFormer on the VoD dataset. We adopt LXL \cite{LXL} supervised by extra LiDAR data as our baseline.}
    \renewcommand\arraystretch{1.1}
    \begin{tabular}{c|c|c|c|c|c}
        \toprule[1.0pt] 
        Baseline & w/ SSI & w/ CVC & w/ IEA & $\text{mAP}_\text{EAA}$ & $\text{mAP}_\text{DC}$ \\
        \midrule
        \checkmark &  &  &  & 56.44 & 73.06 \\
        \checkmark & \checkmark &  &  & 58.69 & 75.24 \\
        \checkmark & \checkmark & \checkmark &  & 59.33 & 76.14 \\
        \checkmark & \checkmark & \checkmark & \checkmark & \textbf{59.94} & \textbf{76.71} \\
        \bottomrule[1.0pt]
    \end{tabular}
    \label{tab:ablation_ALL}
\end{table}

\section{Experiments}
\subsection{Implementation Details}
\textbf{Datesets and Evaluation Metrics.} 
We use the VoD dataset \cite{VoD}, the TJ4DRadset dataset \cite{TJ4D} and the nuScenes dataset \cite{nuscenes} to evaluate our model. The VoD dataset, collected in Delft, contains 5139 training, 1296 validation, and 2247 test frames. The TJ4DRadSet dataset, collected in Suzhou, features challenging conditions such as nighttime, glare, under-bridge scenes, and defocused images, which are difficult for camera-only systems. It includes 7757 frames, 5717 for training and 2040 for testing. The nuScenes dataset is a large-scale benchmark containing data from six cameras, one LiDAR, and five radars. It includes 1000 scenarios, divided into 700 for training, 150 for validation, and 150 for testing.

For the VoD, according to the official recommendations, we use two metrics, \emph{i.e.}, 3D AP under the entire annotated area and 3D AP under the driving corridor. For the TJ4DRadset, 3D AP and BEV AP are evaluated for objects up to 70 m from the radar source. The IoU thresholds are same with the \cite{RCFusion}. For nuScenes, we report nuScenes Detection Score (NDS) and mean Average Precision (mAP).

\textbf{Network Settings and Training details.}
For the VoD dataset, the point cloud range is limited to (0, 51.2) m, ($-$25.6, 25.6) m, and ($-$3, 2.76) m along the $X$-, $Y$-, and $Z$-axes, respectively. We use radar point clouds accumulated over 5 scans as input. For the TJ4DRadSet dataset, the point cloud range is limited to (0, 69.12) m, ($-$39.68, 39.68) m, and ($-$4, 2) m, respectively and single-frame radar point clouds are used as input. For both datasets,
the voxel is set as a cube of size 0.16m. The image is resized to 800×1280 for VoD dataset and 640×800 for TJ4DRadSet dataset, while the number of discretized depth bins is set to 56 for VoD and 72 for TJ4DRadSet. The anchor size and point cloud range for both datasets are kept the same as in \cite{RCFusion}. For the nuScenes dataset, we follow the same setup as RCBEVDet\cite{RCBEVDet}.

We implement our model based on the MMDetection3D \cite{MMDetection3D} framework. The models are trained on 4 NVIDIA GeForce RTX 4090 GPUs with a batch size of 2 per GPU. Our training process consisted of two stages. First, we train the image branch for depth estimation and radar branch for 3D object detection, respectively. The image branch inherites weights from the model pretrained on the COCO\cite{COCO} and KITTI\cite{KITTI} datasets following \cite{RCFusion}, while we train radar branch weights from scratch. Second, we train our model using the weights inherited from the above streams. During the fusion training, we use the AdamW \cite{AdamW} optimizer with an initial learning rate of $1\times 10^{-4}$ and trained the model for 24 epochs for 4D radar based dataset. We adopt image data augmentations including random cropping, random scaling, random flipping, and random rotation, and also adopt BEV data augmentations including random scaling, random flipping, and random rotation. All ablation experiments were conducted on the VoD validation set, utilizing only one-third of the training epochs with an initial learning rate of $3\times 10^{-4}$.

% For the VoD dataset, the point cloud range is limited to (0, 51.2) m, ($-$25.6, 25.6) m, and ($-$3, 2.76) m along the $X$-, $Y$-, and $Z$-axes, respectively. We use radar point clouds accumulated over 5 scans as input. For the TJ4DRadSet dataset, the point cloud range is limited to (0, 69.12) m, ($-$39.68, 39.68) m, and ($-$4, 2) m, respectively and single-frame radar point clouds are used as input. Our training process consisted of two stages. First, we train the image branch for depth estimation and radar branch for 3D object detection, respectively. Second, we train our model using the weights inherited from the above streams. During the fusion training,  All ablation experiments were conducted on the VoD validation set, utilizing only one-third of the training epochs with an initial learning rate of $3\times 10^{-4}$. We implement our model based on the MMDetection3D framework.

\begin{figure*}[t]
    \centering
    \includegraphics[width=0.90\linewidth]{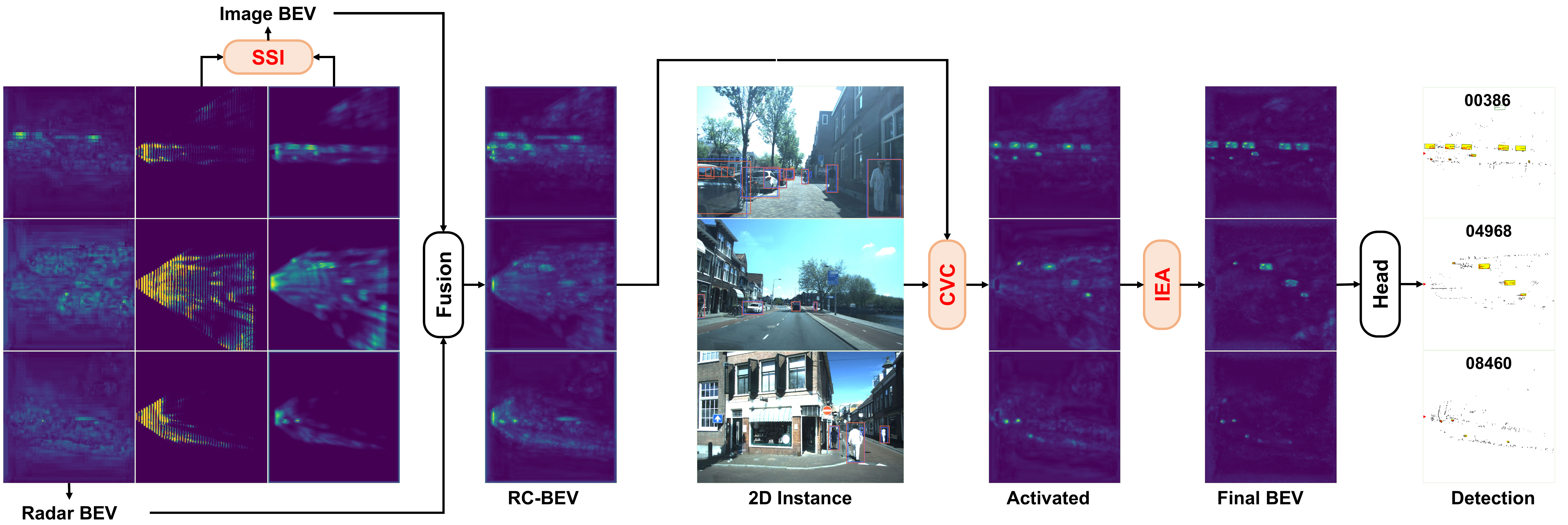}
    \caption{Visualization of the interactions between key components in the SIFormer: Sparse Scene Integration (SSI), Cross-View Correlation (CVC), and Instance Enhance Attention (IEA). The figure shows how feature maps are progressively fused and processed, resulting in the activation of instances and the final detection output. Notably, the CVC mechanism reduces noise from the 4D radar and mitigates blurriness caused by inaccurate view transformations, thereby enhancing the clarity of instance-related regions. Zoom in for better view.}
    \label{fig:features}
\end{figure*}

\begin{table*}[t]
    \belowrulesep=0pt
    \aboverulesep=0pt
    \centering
    \footnotesize
    \caption{Results on robustness setting of modality failure cases on the VoD dataset. FAR denotes the area outside the driving corridor within the entire annotated region. We record the mAP only.}
    \renewcommand\arraystretch{1.1} % 增加表格行距
    \begin{tabular}{c|c|ccc|ccc|ccc}
        \toprule[1.0pt] 
        \multirow{2}{*}{\makebox[0.75cm]{Method}} & 
        \multirow{2}{*}{\makebox[0.75cm]{Modality}} &
        \multicolumn{3}{c|}{4D Radar + Camera} & 
        \multicolumn{3}{c|}{Only Camera} & 
        \multicolumn{3}{c}{Only 4D Radar} \\
        \cmidrule(lr){3-5} \cmidrule(lr){6-8} \cmidrule(lr){9-11}
        & & \makebox[0.75cm]{$\text{mAP}_\text{EAA}$} & \makebox[0.75cm]{$\text{mAP}_\text{DC}$} & \makebox[0.75cm]{$\text{mAP}_\text{FAR}$} 
        & \makebox[0.75cm]{$\text{mAP}_\text{EAA}$} & \makebox[0.75cm]{$\text{mAP}_\text{DC}$} & \makebox[0.75cm]{$\text{mAP}_\text{FAR}$} 
        & \makebox[0.75cm]{$\text{mAP}_\text{EAA}$} & \makebox[0.75cm]{$\text{mAP}_\text{DC}$} & \makebox[0.75cm]{$\text{mAP}_\text{FAR}$} \\
        \midrule
        ImVoxelNet (WACV 2022) \cite{imvoxelnet} & C & - & - & - & 14.17 & 29.39 & - & - & - & - \\
        RadarPillarNet (IEEE TIM 2023) \cite{RCFusion} & R & - & - & - & - & - & - & \textbf{46.33} & 67.08 & \textbf{31.87} \\
        LXL \cite{LXL} (IEEE TIV 2023) & R+C & 56.44 & 73.06 & 44.08 & 1.77 & 3.19 & 0.38 & 41.69 & 63.29 & 27.48 \\
        SGDet3D (IEEE RAL 2024) \cite{SGDet3D} & R+C & 59.75 & 77.42 & 45.05 & 6.90 & 8.85 & 1.05 & 20.88 & 35.76 & 10.21 \\
        ${\text{SIFormer}^{\dagger}}$ (\textbf{Ours}) & R+C & \textbf{63.32} & \textbf{83.06} & \textbf{48.91} & \textbf{17.22} & \textbf{34.88} & \textbf{7.27} & 46.22 & \textbf{67.42} & 31.41\\
        \bottomrule[1.0pt]
    \end{tabular}
    \label{tab:robustness}
\end{table*}

\begin{table}[t]
    \belowrulesep=0pt
    \aboverulesep=0pt
    \centering
    \footnotesize
    \caption{Ablation study on BEV-level fusion components on the VoD dataset. We adopt LXL \cite{LXL} supervised by extra LiDAR data as our baseline.}
    \renewcommand\arraystretch{1.1}
    \begin{tabular}{c|c|c|c|c}
        \toprule[1.0pt]
        \multirow{2}{*}{\makebox[0.75cm]{Baseline}} & \multicolumn{2}{c|}{SSI} & \multirow{2}{*}{\makebox[0.75cm]{$\text{mAP}_\text{EAA}$}} & \multirow{2}{*}{\makebox[0.75cm]{$\text{mAP}_\text{DC}$}} \\
        \cmidrule(lr){2-3}
        & \makebox[1.95cm]{w/ DGW} & \makebox[1.95cm]{w/ SGW} & & \\
        \midrule
        \checkmark &  &  & 56.44 & 73.06 \\
        \checkmark & \checkmark &  & 57.44 & 74.40 \\
        \checkmark & & \checkmark & 56.98 & 74.22 \\
        \checkmark & \checkmark & \checkmark & \textbf{58.69} & \textbf{75.24} \\
        \bottomrule[1.0pt]
    \end{tabular}
    \label{tab:ablation_bev_level}
\end{table}

\begin{table}[t]
    \belowrulesep=0pt
    \aboverulesep=0pt
    \centering
    \footnotesize
    \caption{Ablation study on perspective-level fusion on the VoD dataset.}
        \renewcommand\arraystretch{1.1} %增加表格行距
    \begin{tabular}{c|c|c|c|c|c|c}
        \toprule[1.0pt] 
        \multirow{2}{*}{\makebox[0.75cm]{Baseline}} & 
        \multicolumn{2}{c|}{CVC} & 
        \multicolumn{2}{c|}{IEA} & 
        \multirow{2}{*}{\makebox[0.75cm]{$\text{mAP}_\text{EAA}$}} &
        \multirow{2}{*}{\makebox[0.75cm]{$\text{mAP}_\text{DC}$}}\\
        \cmidrule(lr){2-3} \cmidrule(lr){4-5} 
        & \makebox[0.75cm]{w/ Det2D} & \makebox[0.75cm]{w/ FDL} & \makebox[0.75cm]{w/ GEM} & \makebox[0.75cm]{w/ SEM} & \\
        \midrule
        \text{\checkmark} & & & & & 58.69 & 75.24\\
        % \text{\checkmark} & & & & \text{\checkmark} & 59.45 & 76.65\\
        \text{\checkmark} & \text{\checkmark} & & & & 58.97 & 75.20\\
        \text{\checkmark} & \text{\checkmark} & \text{\checkmark} & & & 59.33 & 76.14\\
        \text{\checkmark} & \text{\checkmark} & \text{\checkmark} & \text{\checkmark} & & 59.47 &76.47\\
        \text{\checkmark} & \text{\checkmark} & \text{\checkmark} & \text{\checkmark} & \text{\checkmark} & \textbf{59.94} & \textbf{76.71}\\
        \bottomrule[1.0pt]
    \end{tabular}
    \label{tab:ablation perspective-level fusion}
\end{table}

\begin{table}[t]
    \belowrulesep=0pt
    \aboverulesep=0pt
    \centering
    \footnotesize
    \caption{Ablation study on cross view correlation (CVC) of the perspective-level fusion on the VoD dataset.}
        \renewcommand\arraystretch{1.1} %增加表格行距
    \begin{tabular}{c|c|c|c|c|c|c}
        \toprule[1.0pt] 
        \multirow{2}{*}{\makebox[0.75cm]{Baseline}} & 
        \multicolumn{4}{c|}{Feature Disentanglement Learning} & 
        \multirow{2}{*}{\makebox[0.75cm]{$\text{mAP}_\text{EAA}$}} &
        \multirow{2}{*}{\makebox[0.75cm]{$\text{mAP}_\text{DC}$}}\\
        \cmidrule(lr){2-5} 
        & \makebox[0.75cm]{w/ $\mathbf{T_{\text{q}}}$ } & \makebox[0.75cm]{w/ $\mathcal{L}_\text{seg}$} & \makebox[0.75cm]{w/ $\mathcal{L}_\text{neg}$} & \makebox[0.75cm]{w/ back} & \\
        \midrule
        \text{\checkmark} & & & & & 58.97 & 75.36\\
        \text{\checkmark} & \text{\checkmark} & & & & 58.85 & 75.32\\
        \text{\checkmark} & \text{\checkmark} & \text{\checkmark} & & & 59.28& 75.98\\
        \text{\checkmark} & \text{\checkmark} & \text{\checkmark} & \text{\checkmark} & & 59.30 & 76.05\\
        \text{\checkmark} & \text{\checkmark} & \text{\checkmark} & \text{\checkmark} & \text{\checkmark} & \textbf{59.33} & \textbf{76.14}\\
        \bottomrule[1.0pt]
    \end{tabular}
    \label{tab:ablation feature disentangle learning}
\end{table}

\subsection{3D Object Detection Results}

\textbf{Results on 4D Radar Dataset.} 
Table \ref{tab:VoD} and Table \ref{tab:TJ4D} presents the 3D object detection results on the VoD validation set and TJ4DRadSet test set, where our proposed SIFormer consistently outperforms state-of-the-art radar-camera fusion models. The enhanced version ${\text{SIFormer}^{\dagger}}$, which uses LiDAR data for depth supervision, achieves the notably highest results. For cars, it achieves outstanding accuracy, while for pedestrians, which are more challenging due to small size and occlusions, our model still leads on both metrics. However, SIFormer shows relatively weaker results on cyclists, possibly due to visual ambiguity between bicycles and the background or between riders and pedestrians. Compared to the strong baseline LXL \cite{LXL}, ${\text{SIFormer}^{\dagger}}$ brings significant gains of 7.01\% in $\text{mAP}_{\text{EAA}}$ and 10.13\% in $\text{mAP}_{\text{DC}}$, showcasing its ability to leverage both BEV-level and perspective-level fusion for deep cross-modal interaction. Importantly, even without LiDAR supervision, SIFormer remains comparable to ${\text{SGDet3D}^{\dagger}}$, which use LiDAR.

Visualization results on the VoD and TJ4DRadSet dataset are shown in Fig. \ref{fig:Visualization}. In (a) of VoD, our model even detects a distant pedestrian without ground-truth annotation, which we attribute to the perspective-level 2D object detection mechanism. This highlights how perspective-level fusion enhances instance awareness, while BEV-level fusion compensates for limited global perception, ultimately enabling comprehensive detection. Results on TJ4DRadSet illustrate the robustness of SIFormer in the challenging environments of TJ4DRadSet. Additionally, we have included more visualizations of the predictions in Fig. \ref{fig:3Dvisualization}.

\textbf{Results on 3D Radar Dataset.}
In Table \ref{tab:nuscenes}, we report the results of our model on the nuScenes validation set \cite{nuscenes}, which only offers 3D radar. ${\text{SIFormer}^{\dagger}}$ achieves first place in NDS and second place in mAP, showcasing its robust radar-camera fusion capabilities. Our model also adapts well to 3D radar, and its performance significantly surpasses that of approaches relying purely on camera data, highlighting the important role of radar in 3D object detection. Since our method is designed for 4D radar, it achieves a large margin improvement over existing model on 4D radar datasests VoD\cite{VoD} and TJ4DRadSet\cite{TJ4D}, while on nuScenes, it only slightly outperforms existing approaches. This difference may be attributed to the disparity between 3D radar and 4D radar data.

\subsection{Robustness Against Sensor Failure.}
We evaluate the robustness of our model under sensor malfunction scenarios, as shown in Table \ref{tab:robustness}. SIFormer consistently outperforms both single-modal and multi-modal baselines under various 4D radar and camera failure conditions. In the monocular-only setting, SIFormer significantly surpasses fusion-based approaches like LXL \cite{LXL} and SGDet3D \cite{SGDet3D}, as well as the vision-only baseline ImVoxelNet \cite{imvoxelnet}, showcasing its effectiveness in critical camera-only scenarios. For radar-only input, SIFormer maintains high performance, whereas SGDet3D suffers due to its complex fusion design, which highlight the robustness of our architecture.

\subsection{Inference Speed.}
As for inference speed, 4D radar single-modality 3D object detection typically achieves higher speeds. SIFormer achieves 6.9 FPS, slightly faster than the baseline LXL \cite{LXL}, while delivering a significant performance boost. Additionally, ${\text{SIFormer}^{\dagger}}$ uses LiDAR data only during training and not during inference, which does not affect FPS.

\begin{table}[t]
    \belowrulesep=0pt
    \aboverulesep=0pt
    \centering
    \footnotesize
    \caption{Ablation study on depth label used for supervising depth prediction. Experiment is conducted using the complete SIFormer network. Half training epochs.}
        \renewcommand\arraystretch{1.1} %增加表格行距
    \begin{tabular}{c|c|c|c|c|c}
        \toprule[1.0pt]
        \multirow{2}{*}{\makebox[1.00cm]{Setting}} & 
        \multicolumn{3}{c|}{View-of-Delft Dataset} & 
        \multirow{2}{*}{\makebox[0.75cm]{$\text{mAP}_\text{EAA}$}} &
        \multirow{2}{*}{\makebox[0.75cm]{$\text{mAP}_\text{DC}$}}\\
        \cmidrule(lr){2-4} 
        & \makebox[1.00cm]{None} 
        & \makebox[1.00cm]{Radar} 
        & \makebox[1.00cm]{LiDAR} & \\
        \midrule
        SGDet3D & \text{\checkmark} & & & 54.12 & 72.89\\ 
        Ours & \text{\checkmark} & & & 56.05 & 74.34\\ 
        \hline
        SGDet3D & & \text{\checkmark} & & 56.67 & 74.03\\
        Ours & & \text{\checkmark} & & 58.26 & 75.11\\ 
        \hline
        SGDet3D & & & \text{\checkmark} & 58.11 & 76.02\\
        Ours & & & \text{\checkmark} & \textbf{59.94} & \textbf{76.71}\\ 
        \midrule
        \multirow{2}{*}{\makebox[1.00cm]{Setting}} & 
        \multicolumn{3}{c|}{TJ4DRadSet Dataset} & 
        \multirow{2}{*}{\makebox[0.75cm]{$\text{mAP}_\text{3D}$}} &
        \multirow{2}{*}{\makebox[0.75cm]{$\text{mAP}_\text{BEV}$}}\\
        \cmidrule(lr){2-4} 
        & \makebox[1.00cm]{None} 
        & \makebox[1.00cm]{Radar} 
        & \makebox[1.00cm]{LiDAR} & \\
        \midrule
        SGDet3D & \text{\checkmark} & & & 38.11 & 45.58\\ 
        Ours & \text{\checkmark} & & & 39.85 & 46.61 \\ 
        \hline
        SGDet3D & & \text{\checkmark} & & 41.22 & 46.99\\ 
        Ours & & \text{\checkmark} & & 42.59 & 48.18\\ 
        \hline
        SGDet3D & & & \text{\checkmark} & 41.82 & 47.16\\ 
        Ours & & & \text{\checkmark} & \textbf{42.61} & \textbf{49.12}\\ 
    \bottomrule[1.0pt]    
    \end{tabular}
    \label{tab:ablation depth label}
\end{table}

\begin{table}[t]
    \belowrulesep=0pt
    \aboverulesep=0pt
    \centering
    \footnotesize
    \caption{Ablation study on the impact of sparse radar depth and LSS \cite{LSS} participation on detection performance. We adopt LXL \cite{LXL} supervised by extra LiDAR data as our baseline.}
    \renewcommand\arraystretch{1.1}
    \begin{tabular}{c|c|c|c|c|c}
        \toprule[1.0pt]
        \makebox[1.00cm]{Baseline} & \makebox[1.00cm]{w/ depth} & \makebox[1.00cm]{w/ LSS} & \makebox[1.00cm]{w/ SSI} & \makebox[0.75cm]{$\text{mAP}_\text{EAA}$} & \makebox[0.75cm]{$\text{mAP}_\text{DC}$} \\
        \midrule
        \checkmark &  &  &  & 56.44 & 73.06 \\
        \checkmark & \checkmark &  &  & 56.97 &  73.43  \\
        \checkmark &  & \checkmark &  & 56.22 & 73.24 \\
        \checkmark &  &  & \checkmark & 55.53 & 71.63 \\
        \checkmark & \checkmark & \checkmark &  & 57.44 & 74.40  \\
        \checkmark & \checkmark & \checkmark & \checkmark & \textbf{58.69} & \textbf{75.24} \\
        \bottomrule[1.0pt]
    \end{tabular}
    \label{tab:ablation_radar_LSS}
\end{table}

% \begin{table}[t]
%     \belowrulesep=0pt
%     \aboverulesep=0pt
%     \centering
%     \footnotesize
%     \caption{Ablation study on LiDAR depth supervision, conducted using SIFormer and SGDet3D \cite{SGDet3D} on the VoD dataset \cite{VoD}.}
%     \renewcommand\arraystretch{1.1}
%     \begin{tabular}{c|c|c|c|c}
%         \toprule[1.0pt]
%         \multirow{2}{*}{\makebox[1.75cm]{Module}} & \multicolumn{2}{c|}{$\text{mAP}_\text{EAA}$} & \multicolumn{2}{c}{$\text{mAP}_\text{DC}$} \\
%         \cmidrule(lr){2-3} \cmidrule(lr){4-5}
%         & \makebox[1.20cm]{None} & \makebox[1.20cm]{LiDAR} & \makebox[1.20cm]{None} & \makebox[1.20cm]{LiDAR} \\
%         \midrule
%         SGDet3D \cite{SGDet3D} & 56.95 & \textbf{59.49} & 74.12 & \textbf{77.11} \\
%         Ours & 60.18 & \textbf{63.32}  & 77.27 & \textbf{83.06} \\
%         \bottomrule[1.0pt]
%     \end{tabular}
%     \label{tab:ablation_LiDARsup}
% \end{table}

\begin{table}[t]
    \belowrulesep=0pt
    \aboverulesep=0pt
    \centering
    \footnotesize
    \caption{Comparison with IS-Fusion\cite{IS-Fusion}. We replicated IS-Fusion on a 4D radar-based dataset, simply replacing the detection decoder with our anchor-based head.}
    \renewcommand\arraystretch{1.1}
    \begin{tabular}{c|c|c|c|c}
        \toprule[1.0pt]
        \multirow{2}{*}{\makebox[1.75cm]{Module}} & \multicolumn{2}{c|}{$\text{mAP}_\text{EAA}$} & \multicolumn{2}{c}{$\text{mAP}_\text{DC}$} \\
        \cmidrule(lr){2-3} \cmidrule(lr){4-5}
        & \makebox[1.20cm]{IGF} & \makebox[1.20cm]{CVC} & \makebox[1.20cm]{IGF} & \makebox[1.20cm]{CVC} \\
        \midrule
        HSF & 54.40 & 56.88 & 74.75 & 76.44 \\
        SSI & 58.13 & \textbf{59.98}  & 72.59 & \textbf{77.23} \\
        \bottomrule[1.0pt]
    \end{tabular}
    \label{tab:ablation_ISFusion}
\end{table}

\begin{table}[t]
    \belowrulesep=0pt
    \aboverulesep=0pt
    \centering
    \footnotesize
    \caption{Comparison with LXL\cite{LXL} under the disturbance to the calibration matrix.}
    \renewcommand\arraystretch{1.1}
    \begin{tabular}{c|c|c|c|c}
        \toprule[1.0pt]
        \multirow{2}{*}{\makebox[1.75cm]{Disturbance}} & \multicolumn{2}{c|}{$\text{mAP}_\text{EAA}$} & \multicolumn{2}{c}{$\text{mAP}_\text{DC}$} \\
        \cmidrule(lr){2-3} \cmidrule(lr){4-5}
        & \makebox[1.2cm]{LXL} & \makebox[1.2cm]{SIFormer} & \makebox[1.2cm]{LXL} & \makebox[1.2cm]{SIFormer} \\
        \midrule
        None & 56.42 & \textbf{63.32} & 73.03 & \textbf{83.06} \\
        $\pm \text{2}^\circ, \pm \text{0.2m}$ & 56.26 & \textbf{63.07} & 73.16 & \textbf{83.00} \\
        $\pm \text{5}^\circ, \pm \text{0.5m}$ & 50.25 & \textbf{56.50} & 72.07 & \textbf{78.99} \\
        $\pm \text{10}^\circ, \pm \text{1.0m}$ & 39.47 & \textbf{45.55} & \textbf{61.29} & 58.23 \\
        $\pm \text{20}^\circ, \pm \text{1.5m}$ & 31.45 & \textbf{37.45} & 50.85 & \textbf{51.16}\\
        \bottomrule[1.0pt]
    \end{tabular}
    \label{tab:ablation_calibration}
\end{table}

\begin{table}[t]
    \belowrulesep=0pt
    \aboverulesep=0pt
    \centering
    \footnotesize
    \caption{Ablation on depth-bin selection with half training epochs.}
    \renewcommand\arraystretch{1.1}
    \begin{tabular}{c|c|c|c|c}
        \toprule[1.0pt]
        \multirow{2}{*}{\makebox[1.75cm]{Percentage}} 
        & \multicolumn{2}{c|}{View-of-Delft Dataset} 
        & \multicolumn{2}{c}{TJ4DRadSet Dataset} \\
        \cmidrule(lr){2-3} \cmidrule(lr){4-5}
        & \makebox[1.2cm]{$\text{mAP}_{\text{EAA}}$} 
        & \makebox[1.2cm]{$\text{mAP}_{\text{DC}}$}
        & \makebox[1.2cm]{$\text{mAP}_\text{3D}$}
        & \makebox[1.2cm]{$\text{mAP}_{\text{BEV}}$} \\
        \midrule
        None & 58.11 & 75.22 & 42.12 & 45.56 \\
        25\%  & \textbf{59.94} & 76.71 & \textbf{42.99} & \textbf{47.01} \\
        50\%  & 59.29 & 76.37 & 42.81 & 46.75 \\
        100\% & 58.74 & \textbf{76.73} & 42.02 & 46.83 \\
        \bottomrule[1.0pt]
    \end{tabular}
    \label{tab:ablation_depthbin}
\end{table}

\subsection{Ablation Study}
We present a series of ablation studies to analyze the contributions of various components in SIFormer, demonstrating their impact on the overall performance. The key components evaluated include Sparse Scene Integration (SSI), Cross-View Correlation (CVC), and Instance Enhance Attention (IEA).
\textbf{Overall Ablation Study.}
Table \ref{tab:ablation_ALL} present the overall ablation study that identifies the contributions of SSI, CVC and IEA to the detection performance. We adopt LXL \cite{LXL} with additional LiDAR data for depth supervision as our baseline. In our instance initialization stage, our SSI boosts results by filtering interference features within the scene. At the instance awareness enhancement stage, CVC bridges perspective view and bird’s-eye view feature to activate instance-related regions, enabling deep scene-instance interaction. Finally, IEA aggregates image semantics and radar geometry from multi-modality via transformer, further improving performance. Fig. \ref{fig:features} presents the features evolution procedure, where our components progressively enhance instance-awareness within the scene.

\textbf{Ablation on Instance Initialization stage.}
We investigate the effectiveness through detailed feature filtering mechanism of SSI, as shown in Table \ref{tab:ablation_bev_level}. SSI includes a depth-guided localization feature weighted (DGW) module and a segmentation-guided feature weighted (SGW) module. DGW filters depths during view transformation to reduce interference and mitigate inaccuracies, while SGW reweights context feature by filtering background interference through foreground segmentation. Together, SSI increasing object-background contrast and boost instance-awareness within the scene feature, which help better objects localization.

\textbf{Ablation on Instance Awareness Enhancement.}
We evaluate the effectiveness of perspective-level fusion by analyzing the roles of CVC and IEA. CVC first performs 2D object detection to extract instances, then computes correlations with the BEV feature map using a learnable token. Without correlation, 2D detection acts as an auxiliary loss (Det2D), yielding minor gains (Table \ref{tab:ablation perspective-level fusion}). Incorporating correlation through FDL brings notable improvement by activating potential regions of interest, which are then used as queries in the IEA module. IEA includes a geometry and semantics enhancement module. The GEM enriches geometric features by allowing BEV queries to attend to radar BEV representations, while The SEM refines semantic features using contextual information, both lead to performance improvement.

\textbf{Ablation on Feature Disentangle Learning.}
Table \ref{tab:ablation feature disentangle learning} demonstrates the effectiveness of each component in FDL. We first introduce a learnable token $\mathbf{T}_{\text{q}}$ to aggregate perspective-level features and interact with global scene features. Without supervision, this interaction strengthens cross-view alignment but suffers from ambiguity, resulting in suboptimal performance. Introducing the segmentation loss $\mathcal{L}_{\text{seg}}$ to supervise the correlation map provides explicit guidance, enhancing instance-level aggregation and boosting performance. Specifically, $\mathcal{L}_{\text{seg}}$ highlights object regions on the BEV correlation map, guiding $\mathbf{T}_{\text{q}}$ to focus on relevant perspective information. Additionally, the negative loss $\mathcal{L}_{\text{neg}}$ enforces discriminative similarity vectors by aligning BEV features with ground-truth occupied regions. A background counterpart further improves models by applying both supervision to background regions.

\textbf{Ablation on Extra LiDAR Data.}
As shown in Table \ref{tab:ablation depth label}, we investigate the impact of depth supervision on the full SIFormer network. Considering LiDAR as auxiliary input, we evaluate performance on both the VoD and TJ4DRadSet datasets to assess model performance and the role of depth labels. The supervision sources used in our method follow the standard practice adopted in recent works such as SGDet3D \cite{SGDet3D}. Under the setting of weak radar geometry, LiDAR depth supervision can indeed improve performance for both SGDet3D and our model, where the results demonstrate that incorporating LiDAR supervision yields the best performance. Notably, SIFormer also achieves state-of-the-art results without LiDAR depth supervision, further confirming that the effectiveness of our method does not rely on supervision unavailable to competing approaches.

\textbf{Ablation on Radar Depth and LSS.}
To further analyze the components in the instance initialization stage, we conduct an ablation study on the impact of sparse radar depth and LSS participation on detection performance, as shown in Table \ref{tab:ablation_radar_LSS}. When radar depth fusion is excluded and only SSI filters background interference, performance drops due to inaccurate depth estimation. Combining LSS\cite{LSS} with depth significantly improves performance, and further inclusion of SSI reduces interference and mitigates inaccuracies, achieving the best results. Using LSS alone does not improve performance, as the sampling-based view transformation approach of LXL is more advanced \cite{SimpleBEV}. In summary, radar depth provides geometry, while LSS fully utilizes nearby semantics. The combination of both with SSI yields the best performance. 

\textbf{Comparison with IS-Fusion.}
We conduct an ablation study on the scene-instance fusion approach by decomposing and combining components from our method and IS-Fusion\cite{IS-Fusion}. We analyze Hierarchical Scene Fusion (HSF) and Instance-Guided Fusion (IGF) in IS-Fusion\cite{IS-Fusion} by combining them with our SSI and CVC. Table \ref{tab:ablation_ISFusion} presents the experimental results on the VoD. In IS-Fusion, HSF relies on LiDAR to sample image semantics from grid cells occupied by the point cloud, heavily depending on LiDAR’s strong geometry. However, sparse 4D radar lacks sufficient prior information, leading to inferior performance when both HSF and IGF from IS-Fusion are used. Replacing IGF with CVC, which uses BEV-only to extract instance information, introduces cross-view perspective information, compensating for the limited scene perception in HSF and improving performance. On the other hand, combining SSI with IGF results in a slight performance drop in the driving corridor but significantly enhances overall scene detection, as SSI provides a denser scene understanding compared to HSF, while maintaining instance perception. Finally, combining SSI and CVC achieves the best by constructing an comprehensive scene and leveraging cross view correlation to enhance instance awareness.

\textbf{Robustness Evaluation under Calibration Disturbance.}
To assess the robustness of our method under realistic conditions, we introduced different levels of calibration errors between the image and radar sensors. These errors simulate disturbances in the calibration matrix, which are common in practical scenarios. Table \ref{tab:ablation_calibration} shows the performance of our SIFormer compared to LXL\cite{LXL} under various levels of disturbance. As the disturbance increases, performance of both model decreases, but SIFormer consistently outperforms LXL in both $\text{mAP}_\text{EAA}$ and $\text{mAP}_\text{DC}$. For example, with no disturbance, SIFormer achieves a significant advantage over LXL. Even with large disturbance ($\pm \text{20}^\circ$, $\pm \text{1.5m}$), SIFormer maintains a $\text{mAP}_\text{EAA}$ of 37.45, while LXL drops to 31.45. These results demonstrate that SIFormer is more robust to calibration errors, making it better suited for real-world applications where sensor calibration may not be perfect.

\textbf{Hyperparameter Study on Depth-bin Selection.}
To evaluate the influence of depth-bin selection, we conducted a hyperparameter study in which different percentages of depth bins were retained. As summarized in Table \ref{tab:ablation_depthbin}, keeping the top 25\% depth bins yields the best overall performance. This setting effectively suppresses irrelevant or noisy depth responses while preserving the most reliable geometric cues, thereby enhancing the robustness of the depth-guided initialization stage. Retaining fewer bins leads to insufficient geometric guidance, while keeping more bins reintroduces background noise. These results validate the necessity of selecting an appropriate proportion of depth bins and support our design choice of using the top 25\%.

\section{Conclusions}
In this work, we have proposed SIFormer, a scene-instance aware transformer for 3D object detection using 4D radar and camera. SIFormer enhances instance awareness within the scene by filtering interference features during view transformation and employing cross-view correlation that effectively leverages 2D detection results from the perspective view. Additionally, the instance enhancement transformer is designed to further improves robustness and performance. As a result, our model effectively combines both perspective-level and BEV-level paradigms through a cross-view mechanism to reveal and establish the relationship between instances and the scene, which addresses the inherent limitation of weak geometry in 4D radar data. Extensive experimental results demonstrate that SIFormer achieves state-of-the-art performance on the public View-of-Delft and TJ4DRadSet datasets.

\emph{Limitation.} 
Despite its strong performance, SIFormer is limited by its inference speed and the absence of temporal modeling. In future work, we will explore lightweight variants through knowledge distillation and pruning, and further incorporate temporal modeling to improve both efficiency and detection robustness.
\bibliographystyle{IEEEtran}
\bibliography{ref}

\begin{IEEEbiography}[{\includegraphics[width=1in,height=1.25in,clip,keepaspectratio]{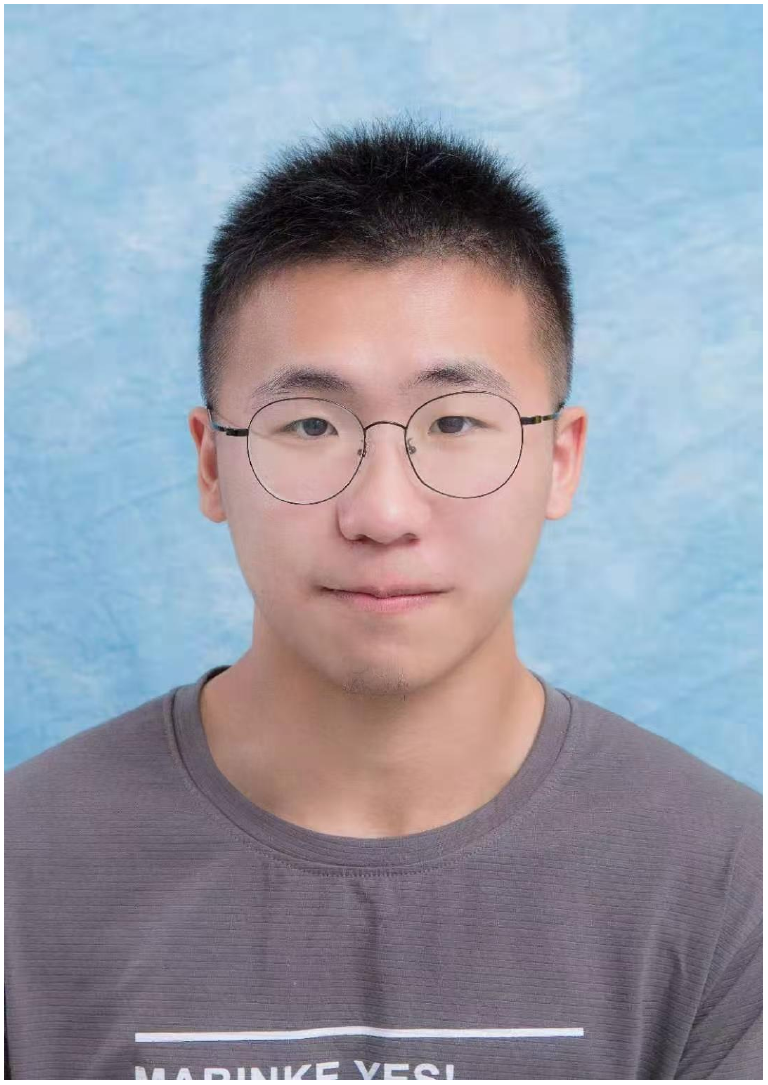}}]{Xiaokai Bai} received his B.Eng. degree from Zhejiang University in 2023. He is currently pursuing the Ph.D. degree with the College of Information Science and Electronic Engineering, Zhejiang University, China. He serves as a peer reviewer for journals
and conferences including TITS, TGRS, RAL, ICRA, IROS and AAAI. His research interests are 4D radar, 3D object detection, 3D collaborative perception and 3D reconstruction.
\end{IEEEbiography}
\vspace{-1cm}

\begin{IEEEbiography}[{\includegraphics[width=1in,height=1.25in,clip,keepaspectratio]{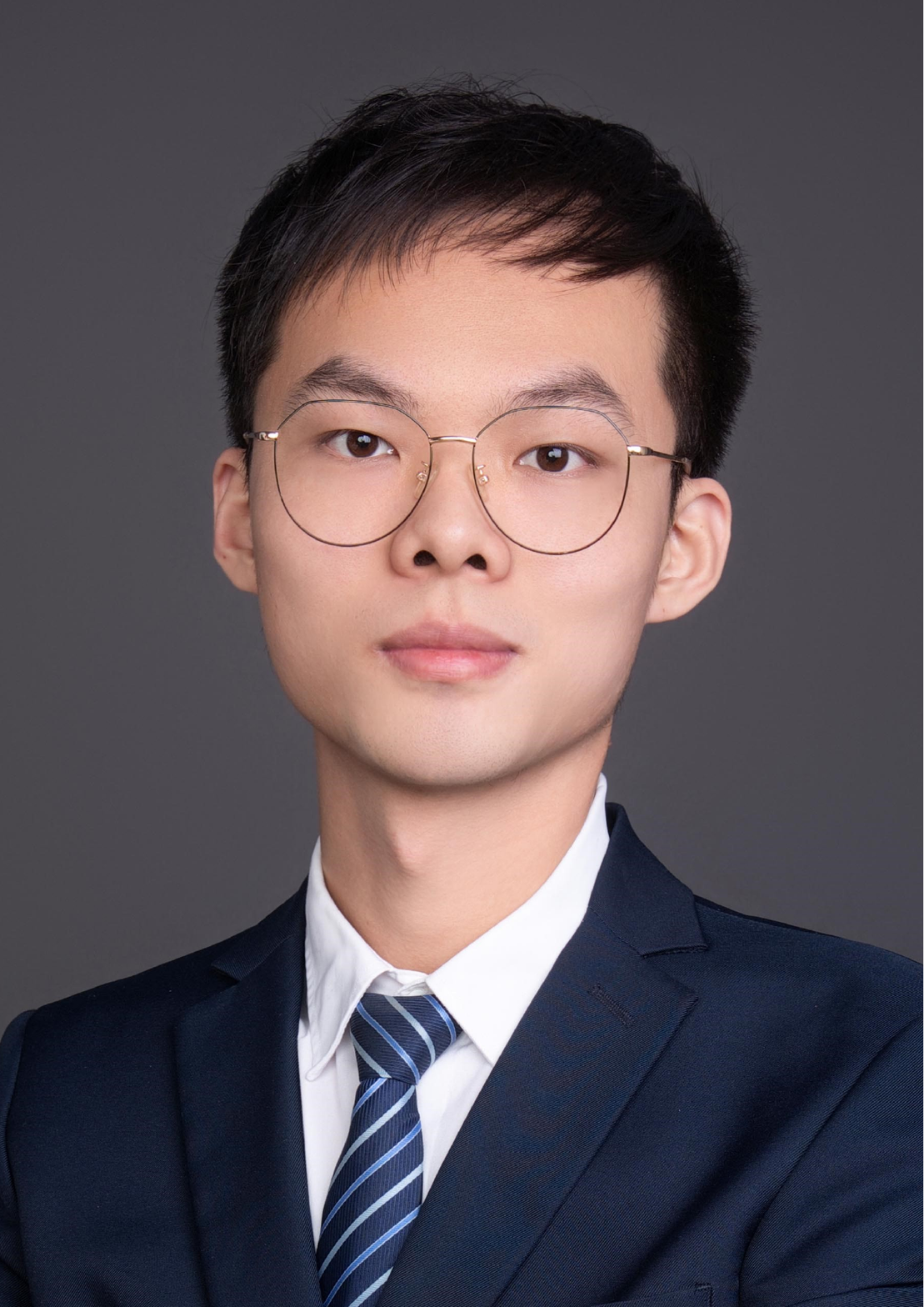}}]{Lianqing Zheng} received his B.S. degree in vehicle engineering from Harbin Institute of Technology, Weihai, China, in 2019. He is currently pursuing a Ph.D. in vehicle engineering at the School of Automotive Studies, Tongji University, Shanghai, China. He serves as a peer reviewer for journals and conferences including \textit{IEEE T-ITS}, \textit{T-IV}, \textit{RA-L}, and \textit{ICRA}. His research interests include data closed-loop systems, multimodal fusion, 4D radar perception, and visual-language models.
\end{IEEEbiography}
\vspace{-1cm}

\begin{IEEEbiography}[{\includegraphics[width=1in,height=1.25in,clip,keepaspectratio]{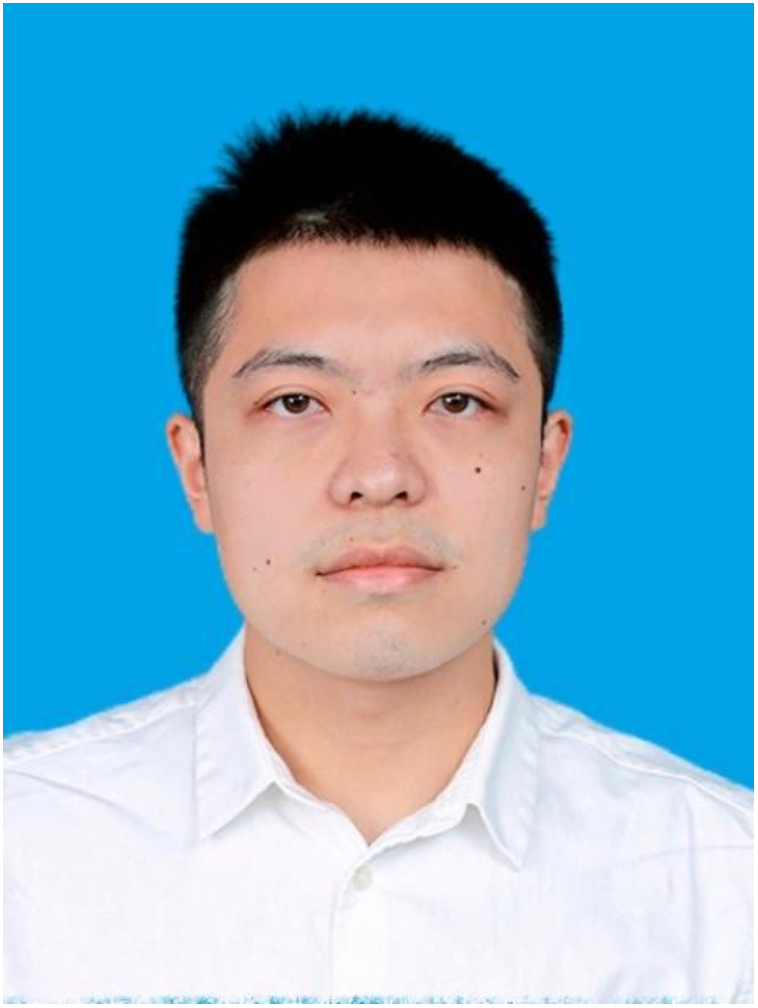}}]{Siyuan Cao} received his B.Eng. degree in electronic information engineering from Tianjin University in 2016, and Ph.D. degree in electronic science and technology from Zhejiang University in 2022. He is currently an assistant researcher in Ningbo Innovation Center, Zhejiang University, China. His research interests are multispectral/multimodal image registration, homography estimation, place recognition and image processing.
\end{IEEEbiography}
\vspace{-1cm}

\begin{IEEEbiography}[{\includegraphics[width=1in,height=1.25in,clip,keepaspectratio]{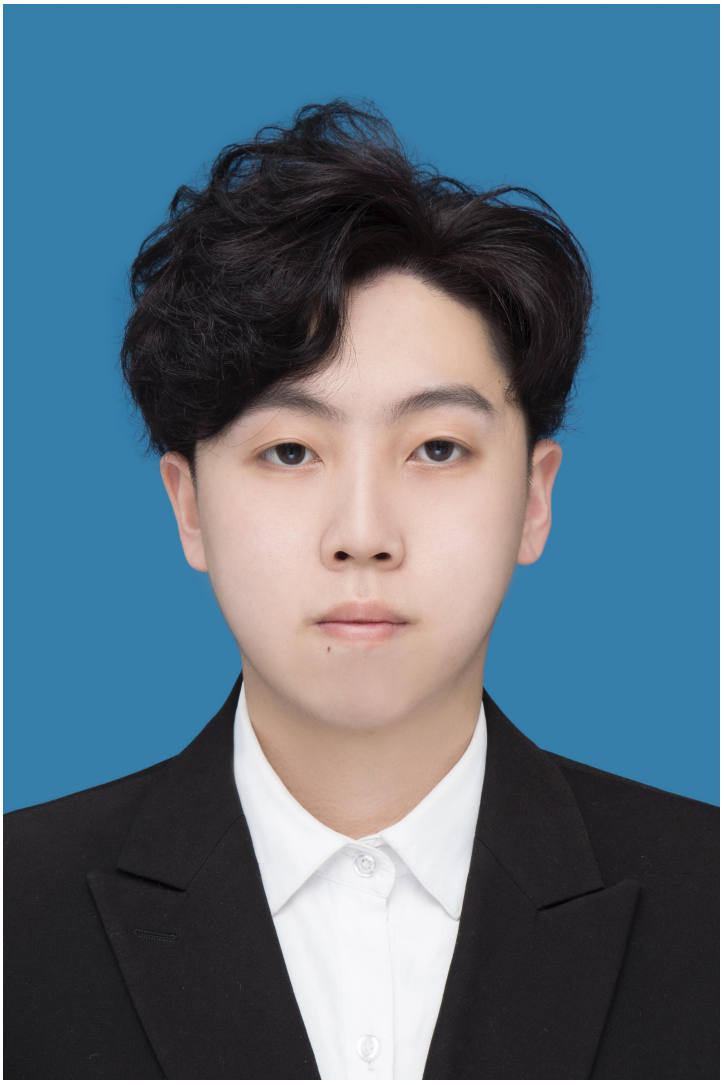}}]{Xiaohan Zhang} received the B.Eng. degrees from Beijing Jiaotong University, China, in 2022. He is currently pursuing the Ph.D. degree with the College of Information Science and Electronic Engineering, Zhejiang University. His research interests are object detection, object geo-localization, and image processing.
\end{IEEEbiography}
\vspace{-1.10cm}

\begin{IEEEbiography}[{\includegraphics[width=1in,height=1.25in,clip,keepaspectratio]{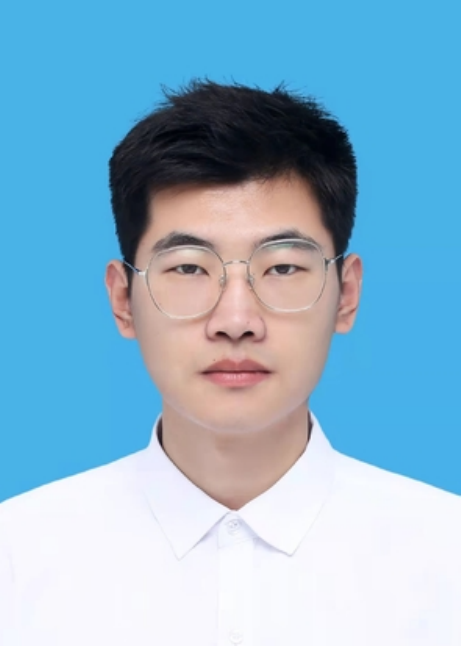}}]{Zhe Wu} received his B.E. degree from Xidian University in 2019 and M.S. degree from the University of Edinburgh in 2020. He is currently pursuing a Ph.D. degree with the College of Information Science and Electronic Engineering, Zhejiang University. His research interests include image restoration and enhancement, computer vision, and deep learning.
\end{IEEEbiography}
\vspace{-1cm}

\begin{IEEEbiography}[{\includegraphics[width=1in,height=1.25in,clip,keepaspectratio]{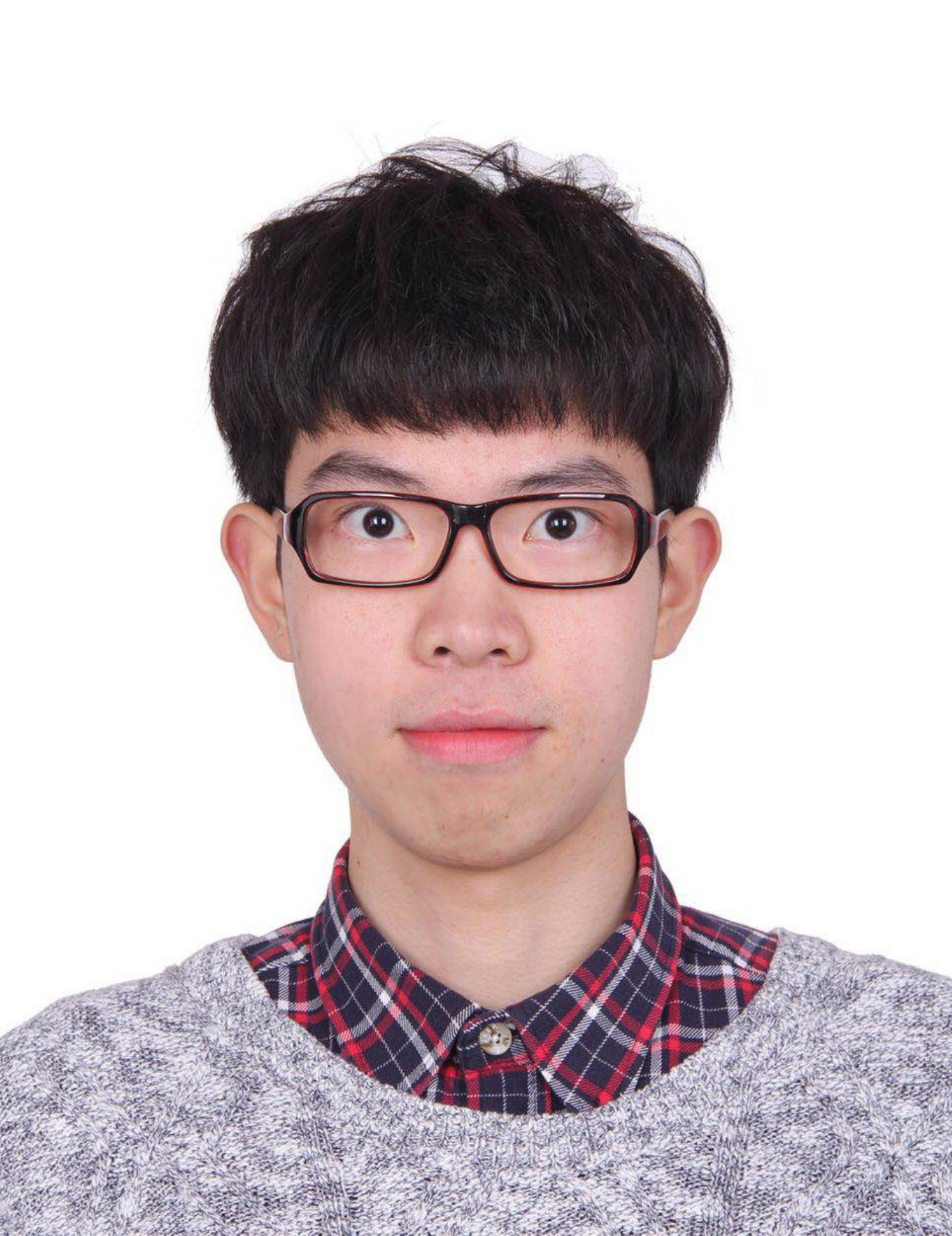}}]{Beinan Yu} received his B.E. degree from Zhejiang University in 2019, and received his Ph.D. degree from Zhejiang University in 2024. He is currently pursuing post doctor work with the College of Computer Science and Technology, Zhejiang University, China. His research interests are multispectral/hyperspectral imaging and image processing.
\end{IEEEbiography}
\vspace{-1cm}

\begin{IEEEbiography}[{\includegraphics[width=1in,height=1.25in,clip,keepaspectratio]{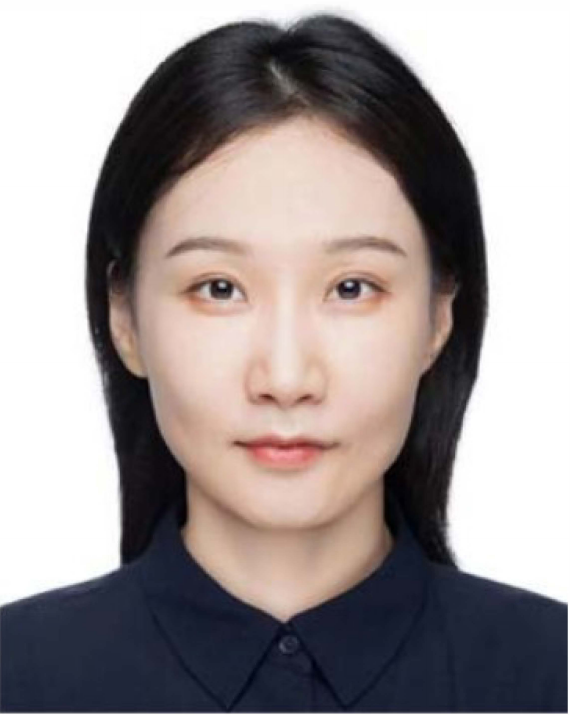}}]{Fang Wang} received the B.Eng. and Ph.D. degrees in design and construction of naval architecture and ocean structure from Harbin Engineering University, Harbin, China, in 2007 and 2012, respectively. She is currently an Associate Professor with School of Information and Electrical Engineering, Hangzhou City University, Hangzhou, China. Her current re-search interests include the autonomous control of unmanned systems and urban air traffic.
\end{IEEEbiography}
\vspace{-1cm}

\begin{IEEEbiography}[{\includegraphics[width=1in,height=1.25in,clip,keepaspectratio]{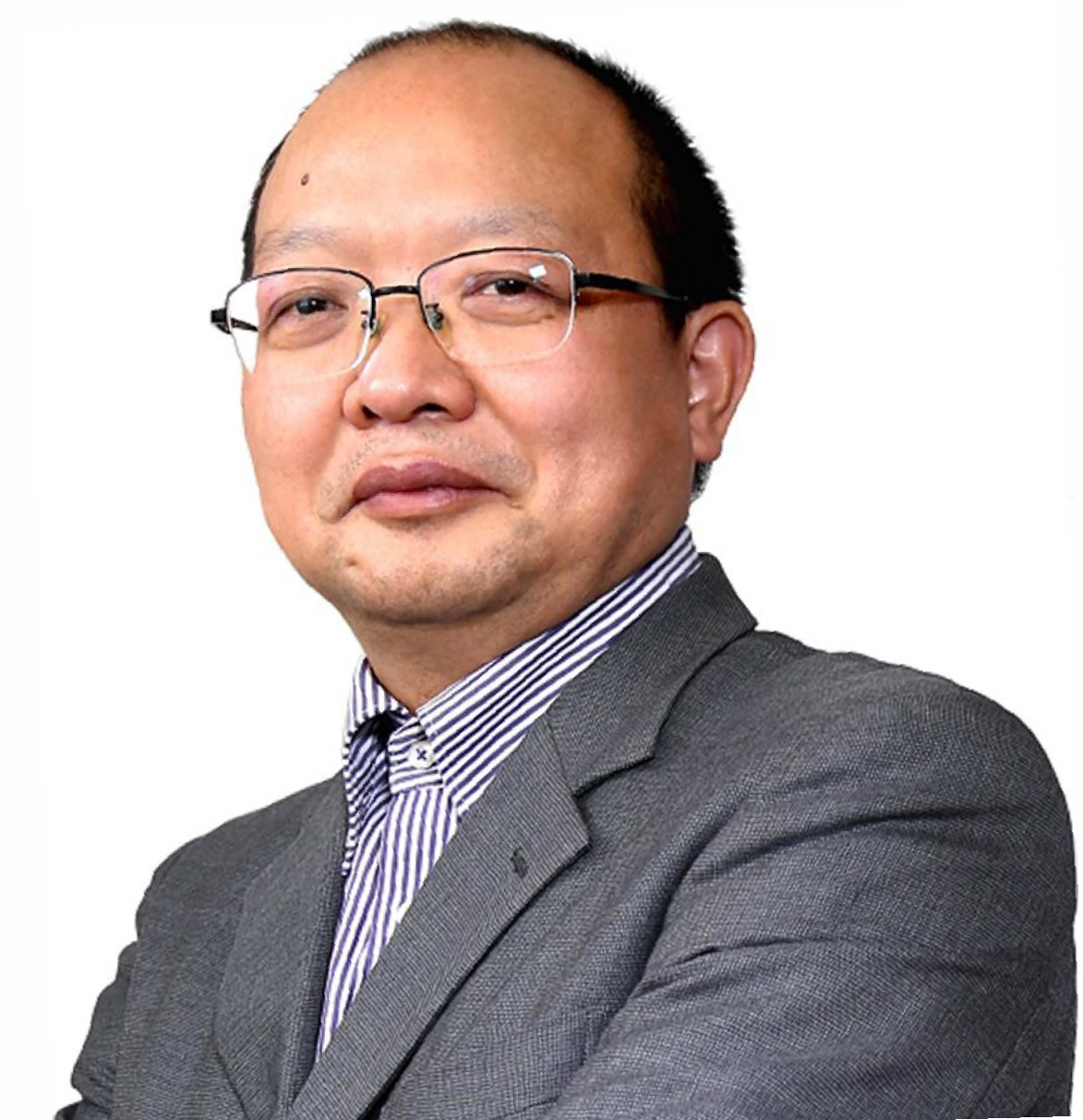}}]{Jie Bai} received the B.S. degree in instrument science and technology from Harbin Institute of Technology, Harbin, China, in 1987. He received the M.S. and Ph.D. degrees in mechanical engineering from Hiroshima University, Japan, in 1992 and 1995, respectively. He is currently a Foreign Fellow of the Russian Academy of Engineering and a Professor with the School of Information and Electrical Engineering, Hangzhou City University, Hangzhou, China. He is the Vice-Chairman of the SAE International Conference Organizing Committee and the Chairman of the Technical Expert Committee. He is also the Secretary General of the Intelligent Transportation Branch of the China Automotive Engineering Institute. His current research interests include intelligent automotive environment perception and decision-making, multi-sensor data fusion, deep learning, and signal processing.
\end{IEEEbiography}
\vspace{-1cm}

\begin{IEEEbiography}[{\includegraphics[width=1in,height=1.25in,clip,keepaspectratio]{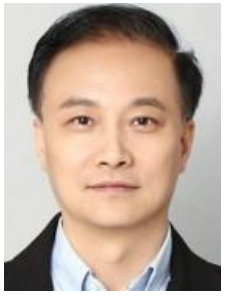}}]{Huiliang Shen}(Senior Member, IEEE) received his B.Eng. and Ph.D. degrees in Electronic Engineering from Zhejiang University, Hangzhou, China, in 1996 and 2002, respectively. He was a Research Associate and Research Fellow with The Hong Kong Polytechnic University from 2001 to 2005. He is currently a Professor with the College of Information Science and Electronic Engineering, Zhejiang University, Hangzhou, China. His research interests are multispectral imaging, image processing, computer vision, deep learning, and machine learning.
\end{IEEEbiography}
\vspace{-1cm}

\vfill

\end{document}